\documentclass[journal]{IEEEtran}
\usepackage{ifpdf}

\usepackage{cite}
\usepackage{hyperref}

\ifCLASSINFOpdf
  \usepackage[pdftex]{graphicx}
  \DeclareGraphicsExtensions{.pdf,.jpeg,.png}
\else
\fi
\usepackage{amsmath}
\usepackage{amsfonts}
\def\a{\mathbf{a}}
\def\x{\mathbf{x}}
\def\y{\mathbf{y}}
\def\z{\mathbf{z}}
\def\v{\mathbf{v}}

\usepackage[ruled,vlined]{algorithm2e}

\usepackage{array}

\ifCLASSOPTIONcompsoc
 \usepackage[caption=false,font=normalsize,labelfont=sf,textfont=sf]{subfig}
\else
 \usepackage[caption=false,font=footnotesize]{subfig}
\fi
\usepackage{booktabs}
\usepackage{multirow}
\begin{document}
%
\title{Discriminative Radial Domain Adaptation}
%
%
%

\author{Zenan Huang,
        Jun Wen,~\IEEEmembership{Member,~IEEE,}
        Siheng Chen,~\IEEEmembership{Member,~IEEE,}
        Linchao Zhu,~\IEEEmembership{Member,~IEEE,}\\
        and Nenggan Zheng,~\IEEEmembership{Senior Member,~IEEE}
\thanks{Zenan Huang is with the Qiushi Academy for Advanced Studies, Zhejiang University, Hangzhou, Zhejiang 310007, China and also with the College of Computer Science and Technology, Zhejiang University, Hangzhou, Zhejiang 310007, China. (e-mail: lccurious@zju.edu.cn).}
\thanks{Jun Wen is with the Department of Biomedical Informatics, Harvard Medical School, Boston, MA, 02115, USA. (e-mail: jun\_wen@hms.harvard.edu).}
\thanks{Siheng Chen is with Cooperative Medianet Innovation Center, Shanghai Jiao Tong University, Shanghai, 200240, China, and also with Shanghai AI Laboratory, Shanghai, 200240, China. (e-mail:sihengc@sjut.edu.cn).}
\thanks{Linchao Zhu is with College of Computer Science and Technology, Zhejiang University, Hangzhou, China. (e-mail: zhulinchao@zju.edu.cn).}
\thanks{Nenggan Zheng is with the Qiushi Academy for Advanced Studies, Zhejiang University, Hangzhou, Zhejiang, 310007, China, also with Collaborative Innovation Center for Artificial Intelligence by MOE and Zhejiang Provincial Government (ZJU) and Zhejiang Lab, Hangzhou, Zhejiang, 311121, China (e-mail: zng@cs.zju.edu.cn).}
\thanks{Nenggan Zheng is the corresponding author.}
}

%
%

\markboth{IEEE TRANSACTIONS ON IMAGE PROCESSING}%
{Shell \MakeLowercase{\textit{et al.}}: Bare Demo of IEEEtran.cls for IEEE Journals}
%



\maketitle

\begin{abstract}
Domain adaptation methods reduce domain shift typically by learning domain-invariant features.
Most existing methods are built on distribution matching, \emph{e.g.}, adversarial domain adaptation, which tends to corrupt feature discriminability.
In this paper, we propose Discriminative Radial Domain Adaptation (DRDA) which bridges source and target domains via a shared radial structure. It's motivated by the observation that as the model is trained to be progressively discriminative, features of different categories expand outwards in different directions, forming a radial structure. We show that transferring such an inherently discriminative structure would enable to enhance feature transferability and discriminability simultaneously. Specifically, we represent each domain with a global anchor and each category a local anchor to form a radial structure and reduce domain shift via structure matching.  It consists of two parts, namely isometric transformation to align the structure globally and local refinement to match each category. To enhance the discriminability of the structure, we further encourage samples to cluster close to the corresponding local anchors based on optimal-transport assignment. Extensively experimenting on multiple benchmarks, our method is shown to consistently outperforms state-of-the-art approaches on varied tasks, including the typical unsupervised domain adaptation, multi-source domain adaptation, domain-agnostic learning, and domain generalization.

\end{abstract}


%
\IEEEpeerreviewmaketitle

\section{Introduction}
%
%
%
%

Machine learning methods generally assume that training and test data come from the same data distribution.
However, such an assumption may not hold in practice, since a model trained on one distribution or one domain may need to be applied to data from another distribution or domain.
Typically, such distribution shifts or domain shifts would cause significant performance drop \cite{panSurveyTransferLearning2010a,ben-davidTheoryLearningDifferent2010a}.
To address this issue, domain adaptation methods are proposed, which aim to generalize the learned knowledge from source domain to target domains. 

\begin{figure}
    \centering
    \includegraphics[width=\linewidth]{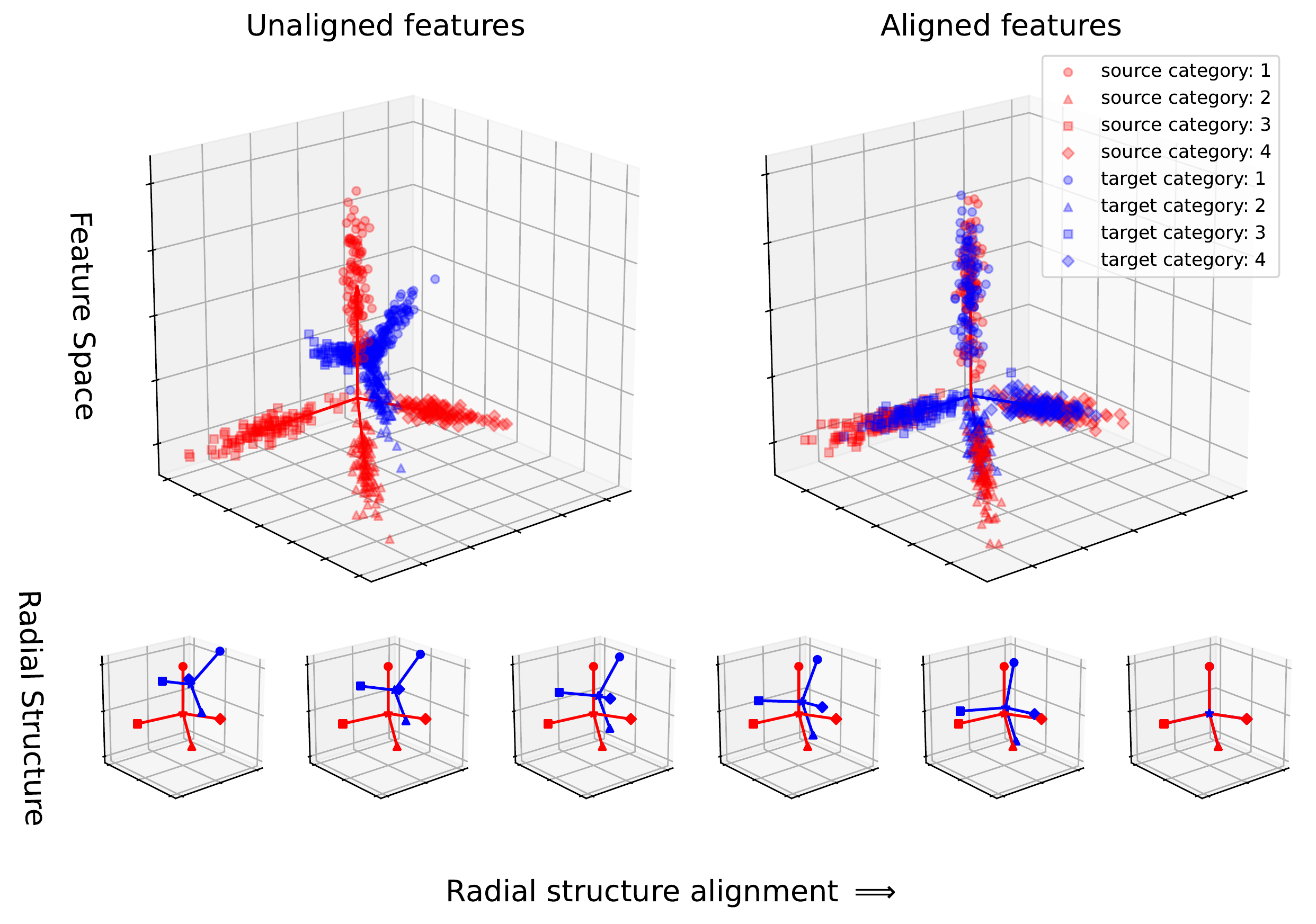}
    \caption{Illustration of the proposed method which represents each domain using a radial structure and reduce domain shift via structure matching (source: red; target: blue; best viewed in color).}
    \label{fig:ball_vis_inituition}
\end{figure}

Domain adaptation methods reduce domain shift typically by learning domain-invariant representations \cite{ben-davidTheoryLearningDifferent2010a}.
Previously, shallow features from both the source and target domains are mapped into a shared subspace \cite{gongConnectingDotsLandmarks2013a}. 
With the success of deep learning, domain-invariant features are learned using deep neural networks \cite{oquabLearningTransferringMidlevel2014}, upon which various domain discrepancy measures are proposed, \emph{e.g.}, Maximum Mean Discrepancy (MMD) \cite{grettonKernelTwoSampleTest2012}, second-order correlations \cite{sunDeepCORALCorrelation2016}, and moments \cite{DBLP:conf/iclr/ZellingerGLNS17}.
Recently, adversarial domain adaptation methods \cite{Tzeng_2017_CVPR,ganinDomainAdversarialTrainingNeural2016} have achieved excellent performances and became the most popular approach by training an additional discriminator network to distinguish the features from different domains \cite{longLearningTransferableFeatures2015,longUnsupervisedDomainAdaptation2016,longConditionalAdversarialDomain2018a,chenTransferabilityVsDiscriminability2019}.
Domain-invariant features are expected to be learned by training the feature extractor to produce features that are indistinguishable by the discriminator.

Though the prevalent adversarial domain adaptation has shown success in many areas, there are still two limitations. 
Firstly, the \emph{minmax} game of adversarial learning is notoriously known to be difficult to optimize, requiring lots of training tricks \cite{aroraGeneralizationEquilibriumGenerative2017}.
When the domain gap is large or the data distribution is complicated with multi modes, these models tend to collapse with false feature alignment \cite{tangDiscriminativeAdversarialDomain2020,kurmiLookingBackLabels2019}, especially when trained from scratch.
Secondly, adversarial training is shown to damage the learned feature discriminability \cite{chenTransferabilityVsDiscriminability2019,Xu_2019_ICCV}.
While shown to be alleviated by either balancing the feature singular values \cite{chenTransferabilityVsDiscriminability2019} or lifting feature norms \cite{Xu_2019_ICCV}, such a discriminability corruption still persists because of the conflicts between transferability and discriminablity which tends to be biased by source labels and with weaker transferability.  
One more promising approach is to learn a discriminative structure that is inherently transferable across domains.

In this paper, we propose Discriminative Radial Domain Adaptation, that gets rid of the typical adversarial learning and bridges the source and target domains via a shared discriminative radial structure. It's motivated by the observation that the features initially all cluster together and as the model is trained to be more discriminative, features of different categories expand outwards in different directions to be more separated in the feature space, forming a radial structure, as also observed in \cite{sunDeepLearningFace2014,wenDiscriminativeFeatureLearning2016a}. 
We bridge the source and target domains by aligning the radial structures. Specifically, we first build a radial structure for each domain that consists of a global domain anchor, which is the centroid of the domain data, and a set of local category anchors. Brute-force matching tends to twist the radial structure and damage its discriminability. To alleviate this, we decompose the structure matching into two components, namely \emph{global isometric transformation} and \emph{local refinement} as shown in Fig.\ref{fig:ball_vis_inituition}. \emph{Global isometric transformation} aims to align the global shape of the two radial structures by bringing close the domain anchors and rotating the overall radial structure using a Sitefel layer \cite{AbsMahSep2008}. To achieve fine-grained alignment of each category, \emph{local refinement} further matches the angles and norms of local category anchors across domains.

To enhance the discriminablity of the radial-like feature distribution, we encourage local features to cluster close to the corresponding local anchors. This is achieved by first assigning each local feature to the optimal local anchors via optimal transport, which prevents false assignments, and then minimizing a optimal-transport distance. Meanwhile, we enforce a prediction consistency between the radial structure and classifier the prevent conditional shift of the classifier. We observe such a consistency also promotes the radial structure to be more discriminative.

The main contributions of this work can be summarized as follows:

\begin{itemize}
  \item We propose a novel domain adaptation approach, called Discriminative Radial Domain Adaptation, that gets rid of the typical adversarial training and reduces domain shift by matching radial structures that are inherently discriminative.
  
  \item We propose to decompose the alignment of radial structure into global isometric transformation and local anchor refinement to prevent damage to the discriminablity of the radial structure.
  
  \item We enhance the discriminablity of the radial structure by minimizing a optimal-transport distance that optimally assigns each feature to the corresponding local anchors to combat false alignment. Further, we perform a prediction consistency between the radical structure and classifier to alleviate conditional shift. 
  
  \item Extensively experimenting on several benchmark datasets, our method outperforms the state-of-the-art approaches not only on the typical single-to-single unsupervised domain adaptation but also on multi-source domain adaptation, domain-agnostic adaptation and domain generalization.
\end{itemize}

\section{Related Works}
In this part, we first review domain adaptation methods and then introduce discriminative structure learning.

\subsection{Domain Adaptation}

To alleviate the domain shift, typical solutions include minimizing domain discrepancy and learning domain invariant features.
In the context of domain discrepancy minimization, approaches can be classified according to their discrepancy metrics and their ways of extracting features.
Discrepancy metrics include the Proxy $\mathcal{A}$-distance \cite{chenMarginalizedDenoisingAutoencoders2012}, the Kullback-Leibler (KL) divergence \cite{zhuangSupervisedRepresentationLearning2015}, the Maximum Mean Discrepancy \cite{grettonKernelTwoSampleTest2012,tzengDeepDomainConfusion2014}, other higher order statistical moments based distance measures \cite{pengMomentMatchingMultiSource2019}, and Optimal Transport distance \cite{kantorovichTranslocationMasses2006,courtyOptimalTransportDomain2017}. 
Many types of feature extraction have also been considered for domain alignment, including handcrafted features \cite{grettonKernelTwoSampleTest2012}, shallow features at the pixel level \cite{bousmalisUnsupervisedPixelLevelDomain2017}, and bottleneck features of deep neural networks \cite{longLearningTransferableFeatures2015,saitoMaximumClassifierDiscrepancy2018}.
Along with their efficiency in reducing marginal domain discrepancies, these methods were found potentially hinder the learning of feature discriminant information \cite{luoConditionalBuresMetric2021}.
Therefore, recent advances have focused on the discrepancy in conditional distribution by using labels or soft labels. 
Such approaches include conditional variants of MMD, Joint Distribution Optimal Transport (JDOT) \cite{courtyJointDistributionOptimal2017}, Moving Semantic Transfer Network (MSTN) \cite{xieLearningSemanticRepresentations2018a}, Robust Spherical Domain Adaptation (RSDA) \cite{guSphericalSpaceDomain2020}, Category-Level Adversarial Network (CLAN) \cite{luoTakingCloserLook2019}, Enhanced Transport Distance \cite{liEnhancedTransportDistance2020}, Discriminative Manifold Propagation \cite{luoUnsupervisedDomainAdaptation2022}, and Conditional Kernel Bures (CKB) metric \cite{luoConditionalBuresMetric2021}. 
These improvements resulting from conditional alignment are evident; however, the changes in the class prior distribution and the noise of estimated target labels also pose risks of misalignment. 
In order to solve these issues, we propose to simultaneously learn the structure of the source and target distributions and align the two domains based on this structure.
That is inspired by factorized optimal transport \cite{linMakingTransportMore2021}, which highlights the benefits of using low-dimensional structures to align data. 
In our framework, domain adaptation is carried out by aligning these radial structures learned from each domain, without relying on sample-level distribution.

Another approach mainly aims at learning domain invariant features so that the target can share the classifier trained from the labeled source.
An effective method to guarantee features transferability is to train the generator to produce indistinguishable features, which can deceive the domain discriminator as a whole, \emph{i.e.} Domain Adversarial Neural Networks (DANN) \cite{ganinDomainAdversarialTrainingNeural2016} and Adversarial Discriminative Domain Adaptation (ADDA) \cite{Tzeng_2017_CVPR}.
In addition, a number of studies have been published that examine ways to improve training strategies in order to create better transferable features from pixel-level features \cite{hoffmanCyCADACycleConsistentAdversarial2018,xuSelfEnsemblingAttentionNetworks2019,xuAdversarialDomainAdaptation2020} or high-level features \cite{liuTransferableAdversarialTraining2019}.
A further effort on producing transferable features is conditional adversarial training discriminator on features and class prediction jointly \cite{longUnsupervisedDomainAdaptation2016,longConditionalAdversarialDomain2018a}. 
Later, more works focus on disentangling original features into domain invariant and domain-specific parts \cite{ijcai2019-285,pengDomainAgnosticLearning2019}.
Nevertheless, these models seek to intensify feature transferability at the expense of feature discriminability. 
In contrast, DRDA applies domain adaptation based on established discriminative radial structures, so the discriminability of features can be well maintained.

\subsection{Discriminative Structure Learning}

Discriminative learning is aimed at pushing dissimilar features away from each other and enclosing the similar ones to be compact.
Many efforts have been made to minimize intra-class feature distances and maximize inter-class feature distances, such as contrastive loss \cite{sunDeepLearningFace2014} and center loss \cite{wenDiscriminativeFeatureLearning2016a}, originally proposed in face recognition tasks.
Inspired by softmax objective, L-Softmax (large margin softmax) \cite{liuLargemarginSoftmaxLoss2016} is introduced as another extension of enhancing discriminability by lifting angular separability between learned features.
The principle of discriminative learning also enhances the performance of domain adaptation tasks.
Follow the discriminative clustering, entropy minimization is introduced into domain adaptation \cite{shiInformationtheoreticalLearningDiscriminative2012,longConditionalAdversarialDomain2018a} to encourage classifier to produce ideal one-hot predictions are promising method.
More recent studies in adversarial-based domain adaptation have shown that the discriminability of target features can be damaged by adversarial feature alignment.
Based on the observations of close relation between the singular values of learned features and discriminative power, \cite{chenTransferabilityVsDiscriminability2019} correct the degeneration of discriminability by adding the penalties corresponding to these singular values.
Also, \cite{Xu_2019_ICCV} identified the connection between norm values and discriminatory power, and then lift the norm values for target features in order to increase the discriminatory power.

As well as features being discriminative during the learning process, the feature distribution is likely to perform in a particular low-dimensional format.
The whole domain can therefore be well sketched by several clusters rather than using entire samples, allowing for a more robust approach to domain adaptation.
In line with this idea, MSTN \cite{xieLearningSemanticRepresentations2018a} is proposed to align the centroids of each category across domains, which reduces the noise influence of false pseudo labels compared to direct matching distributions.
Prototypical networks \cite{Pan_2019_CVPR} is proposed to learn prototypes of each category region and reduce conditional domain discrepancy by learning similar prototypes across domains.
And \cite{liJointFeatureSelection2016} recognizes the importance of structure, connects the statistical property to geometric structures of data, and integrates feature selection and structure preservation into a unified optimization process.
Moreover, \cite{liuStructurePreservedUnsupervisedDomain2019} considered unsupervised domain adaptation as a clustering problem with missing labels using the structure preserve framework.
Compared to these methods, our approach direct models feature distribution with a radial structure which maintains the intrinsic structure of the data while increasing the feature discriminability.

\section{Discriminative Radial Domain Adaptation}
In this section, we first introduce the construction of the radial structure. Then, we describe a proposed structure alignment strategy which decouple alignment into two independent components, namely global isometric transformation and local anchor refinement.

\subsection{Notations and Overview}
In an unsupervised domain adaptation task, we are given labeled source domain $\mathcal{D}_s=\{(\x^s_i, \y^s_i)\}^{n_s}_{i=1}$ of $n_s$ labeled examples and unlabeled target domain $\mathcal{D}_t=\{\x^t_j\}^{n_t}_{j=1}$ of $n_t$ unlabeled examples.
Our model mainly contains a shared backbone $G(\cdot)$ with parameter $\theta$, a shared classifier $F(\cdot)$ with parameter $\varphi$, and a Stiefel layer $S(\cdot)$ whose parameters $\Delta$ are defined on Stiefel manifold $\mathbf{V}_{k}(\mathbb{R}^{d})=\{\Delta \in \mathbb{R}^{d\times k} | \Delta^{\top}\Delta=\mathbf{I}_{k}\}$.
Let $\z_i^s = G(\x_i^s)$ and $\widehat{\y}_i^s = F(\z_i^s)$ be the feature representation and the estimated label of the $i$-th sample in the source domain, respectively.

With insights that linear classification output probability $p_{ik}\propto \exp(W^{T}_{k}\z_{i}+b)=\exp(\|W_{k}\|\|\z_{i}\|\cos(W_{k},\z_{i})+b)$ supposing importance of feature direction and norm in discrimination, we suggest a radial expansion-like structure for modeling features.
Therefore, our framework is aiming to learn and align radial structures $\mathcal{G}^{s}=\{\a^{s},\mathcal{N}^{s}\}$ and $\mathcal{G}^{t}=\{\a^{t},\mathcal{N}^{t}\}$ from source and target domains, each structure containing a \textbf{global anchor} $\a^{s/t}$ and a set $\mathcal{N}^{s/t}=\{\a^{s/t}_{i}\}^{k}_{i=1}$ of $k$ \textbf{local anchors} $\a^{s/t}_{i}\in\mathbb{R}^{d}$.
From an intuitive viewpoint, a radial structure in latent space can be understood as a structure with a group of arrows that point from a global anchor to local anchors.
Thus, for emphasizing the radial chararistic of structures, we also use the egocentric representation version $\mathcal{V}^{s/t}:=\{\v^{s/t}_{i}=(\a^{s/t}_{i}-\a^{s/t})|\a^{s/t}\in \mathcal{N}^{s/t}\}$ of radial structure when comparing the shape differences of the structures.
Finally, domain shifts and class prior differences are then manifested in terms of the isometric transformation and shape differences between two radial structures.
We align the $\mathcal{G}^{s}$ and $\mathcal{G}^{t}$ by reducing isometric transformation to match each other globally in latent space and then refine them into the same shape.
Where the isometric transformation and shape refinement are applied in a non-interfering manner for avoiding negative alignment.
The DRDA approach can be viewed as an alternative optimization strategy that iteratively updates the radial structures $\mathcal{G}^{s},\mathcal{G}^{t}$ to be more representative and aligns radial structures in order to obtain more accurate label predictions in the target domain.

\subsection{Discriminative Radial Structure}

Extraction of radial structure $\mathcal{G}^{s/t}$ includes aggregating \textbf{global anchor} $\mathbf{a}^{s/t}$ and a collection of \textbf{local anchors} $\mathcal{N}^{s/t}$.
We represent the global and local anchors using vectorial embeddings, and iteratively update the anchors and model parameters.

\begin{figure*}
  \centering
  \includegraphics[width=0.95\linewidth]{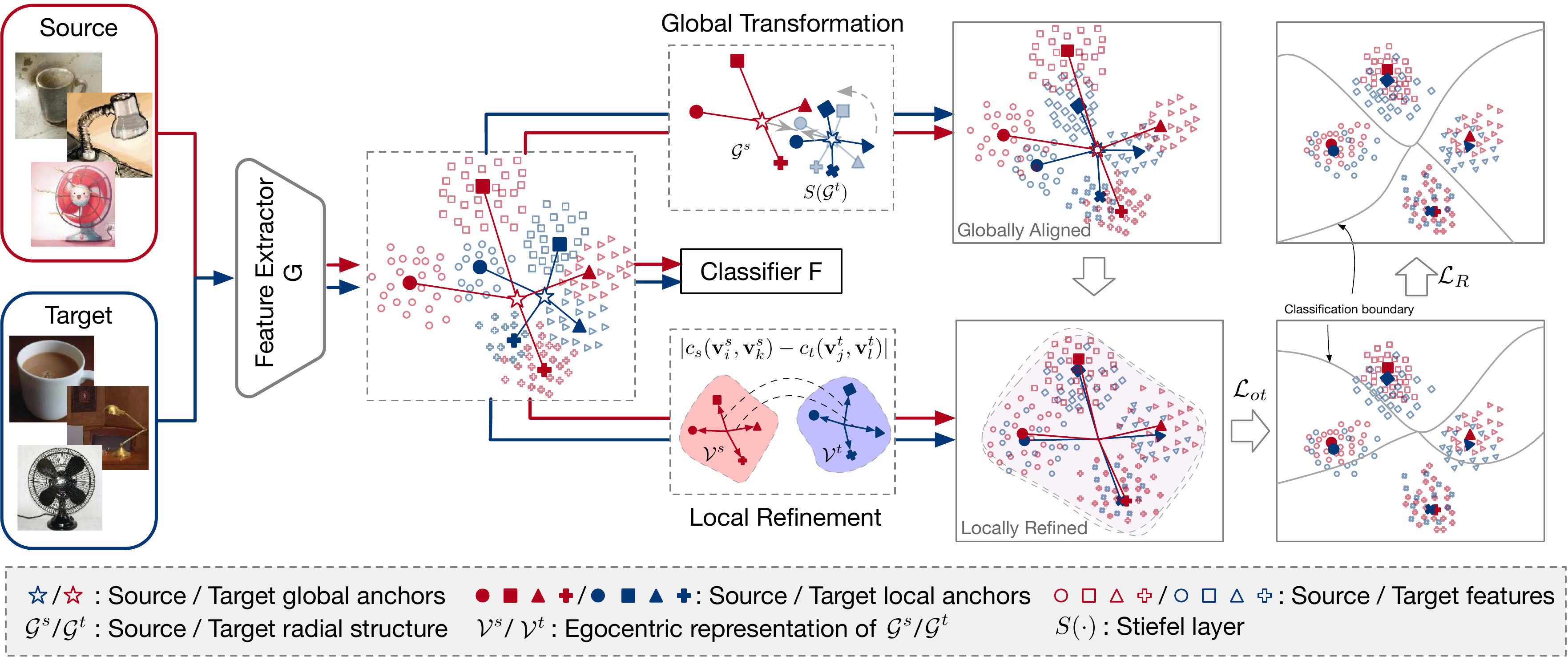}
  \caption{Architecture of Discriminative Radial Domain Adaptation (DRDA). It bridges source and target domains by matching the radial structure which consists of a global domain anchor and a set of local category anchors. DRDA aligns the radial structures across domains via global isometric transformation and local anchor refinement (best viewed in color).}
  \label{fig:hierarchicalmodel}
\end{figure*}

\subsubsection{Global anchors}

For each domain, we define the global anchor as the centroid of overall features extracted by the shared feature extractor $G_{\theta}(\cdot)$.
Formally, the global anchors $\a^{s},\a^{t}$ in the source and target domains are:
\begin{equation}
  \label{eq:domain_anchor}
  \a^{s} = \frac{1}{n_{s}}\sum^{n_{s}}_{i=1} G_{\theta}\left( \x^{s}_{i} \right),\quad \a^{t} = \frac{1}{n_{t}}\sum^{n_{t}}_{j=1} G_{\theta}\left( \x^{t}_{j} \right).
\end{equation}
As an indicator of the mean position of features, global anchor is ideal reference point for contrasting two feature vector under the context of linear classification.
They are also the reference points for comparing the radial structures.
And displacement between global anchors naturally represent the mean feature shift $\mathbb{E}[\mathbf{z}^{s}]-\mathbb{E}[\mathbf{z}^{t}]$.
We then use the distance between $\a^{s}$ and $\a^{t}$ as the global translation distance between two domains.

\subsubsection{Local anchors}
Source and the target radial structures contains $k_{s}$ and $k_{t}$ local anchors, respectively.
Local anchors are located within high-density regions in each domain, each region containing a set of semantically related features.

For the general UDA task, it is straightforward to set $k_{s},k_{t}$ equal to the number of categories to be classified.
Then, the local anchor is equivalent to the centroid of the features with same category.
In labeled source data, such local anchors can be obtained directly from labels, while in the target domain local anchors can be obtained from pseudo-labels:
\begin{equation}
    \label{eq:local_anchors_calculation}
    \a^{s}_{k}=\frac{1}{M_{k}}\sum^{n_{s}}_{i=1}\z_{i}\mathbb{I}[y_{i}=k],\quad \a^{t}_{k}=\frac{1}{M_{k}}\sum^{n_{t}}_{i=1} \z_{i}\mathbb{I}[\hat{y}_{i}=k],
\end{equation}
where $\mathbb{I}[\cdot]$ is indicator function, $M_{k}=\sum\mathbb{I}[y_{i}=k]$ is a normalize constant.

\subsection{Radial Structure Alignment}
We disentangle the structure alignment into two parts, namely global isometric transformation and local anchor refinement, to prevent its corruption to the discriminablity of the radical structure.
It's shown that when $\mathcal{D}_{s}$ and $\mathcal{D}_{t}$ are not aligned, the learned features would be arbitrarily rotated, translated or permuted \cite{titouanSlicedGromovWasserstein2019}.
By disentangling the alignment process into two independent processes, it is hopefully to best prevent false feature alignment.

\subsubsection{Isometric transformation}
We first globally align the shape of the radial structures to reduce the isometric transformation $\tilde{T}(\cdot)$ between source $\mathcal{G}^{s}$ and target $\mathcal{G}^{t}$.
It is equivalent to minimize the isometric transformation $\tilde{T}(\cdot)$ between the source and target, which is defined as:
\begin{equation*}
\tilde{T} := {\arg\min}_{T}\| \mathcal{G}^{s}-T(\mathcal{G}^{t}) \|.
\end{equation*}
We seek to optimize the backbone $G_{\theta}(\cdot)$, thereby making $\tilde{T}(\cdot)$ to be an identical transformation $I(\cdot):I(\mathcal{G})=\mathcal{G}$.

We disentangle the objective into translation and rotation parts in order to optimize feature extractor to achieve isometric alignment.
The translation reduction is performed by minimizing the distance between global anchors among domains as follows:
\begin{equation}
  \label{eq:translation_distance}
  \mathcal{L}_{\rm global} = d(\a^{s}, \a^{t}) = \|\a^{s} - \a^{t}\|_{F},
\end{equation}
where the global distance measures the common distribution difference between source and target.
By applying such global distance minimization, we force two global anchors to align and, consequently, we shift two feature distributions so that they share the same centroid.
In addition, it is intuitively possible to make the entire radial structures (as the structures shown in Fig.\ref{fig:ball_vis_inituition}) on which the feature points lie roughly coincide.

The rotation reduction is accomplished by adding a Stiefel layer $S(\cdot)$ to rotate the target features.
According to a strict definition of the Stiefel manifold $\mathbf{V}_{k}(\mathbb{R}^{d})=\{\Delta \in \mathbb{R}^{d\times k} | \Delta^{\top}\Delta=\mathbf{I}_{k}\}$, the Stiefel layer would perform rotation transform without causing any side effects to the features.
In the backbone networks, we embed the Stiefel layer for target-specific use.
In addition, for emphasizing the radial chararistic, we shift the radial structures $\mathcal{G}^{s},\mathcal{G}^{t}$ with respect to the global anchors, so that they share the same center and are referred to as egocentric version $\mathcal{V}^{s},\mathcal{V}^{t}$, respectively.
Thanks to the benefits of manifold optimization methods \cite{kochurovGeooptRiemannianOptimization2020}, it is easy to optimize the Stiefel layer as following objective:
\begin{equation}
    \Delta^{*} = \mathop{\arg\,\min}_{\mathbf{V}_{k}(\mathbb{R}^{d})}d(\mathcal{V}^{s},S(\mathcal{V}^{t})),
    \label{eq:rotation_distance}
\end{equation}
where the parameter $\Delta$ is optimized with respect to a shape difference metric $d(\cdot, \cdot)$ (induced from local alignment) between the radial structures $\mathcal{G}^{s},\mathcal{G}^{t}$ of the target and the source.
It is noteworthy that the backbone is shared by the source and target, but $\Delta^{*}$ is embed in backbone network for target-specific use.

Optimization on Eq.\eqref{eq:translation_distance} and Eq.\eqref{eq:rotation_distance} enables a coarse alignment among domains, and increases the reliability of target pseudo labels given by the classifier, as discriminative radial structures achieve more and more overlapping as shown in Fig.\ref{fig:hierarchicalmodel}.

\subsubsection{Local refinement}

Global alignment is intended to eliminate isometric discrepancies between the source and the target, whereas local alignment involves refining the two structures to be identical in shape.
In order to avoid the occurrence of a contradiction between global and local alignments, the fine-grained structure difference measured here need to be independent of global alignment.
In light of this, we apply Gromov-Wasserstein (GW) \cite{memoliGromovWassersteinDistances2011} distance to compare the shapes of two radial structures.
According to the definition, GW distance is solely based on intra-space measurements, it has many desirable properties, especially in terms of invariances.
And it terms out the invariant of translations, permutations, and rotations when Euclidean distance is used for intra-space measurement.
Accordingly, whenever two structures $\mathcal{G}^{s}$ and $\mathcal{G}^{t}$ are shifted with different offsets or rotations, GW distance only determines the shape difference between them.
For emphasizing the radial chararistic differences of the source and target structures we use the egocentric representation $\mathcal{V}^{s},\mathcal{V}^{t}$ instead of standard form $\mathcal{G}^{s},\mathcal{G}^{t}$.
Accordingly, the GW distance is defined as follows:
\begin{equation}
    GW^{2}_{2}(c_{s},c_{t},\mu,\nu)=\min_{\pi\in\Pi(\mathcal{V}^{s},\mathcal{V}^{t})}J(c_{s},c_{t},\pi),
\end{equation}
where
\begin{equation*}
    J(c_{s},c_{t},\pi)=\sum_{i,j,k,l}|c_{s}(\v^{s}_{i},\v^{s}_{k})-c_{t}(\v^{t}_{j},\v^{t}_{l})|^{2}\pi_{i,j}\pi_{k,l},
\end{equation*}
where, $\mu=\sum^{k_{s}}_{i=1}\delta_{\a^{s}_{i}}$, $\nu=\sum^{k_{t}}_{i=1}\delta_{\a^{t}_{i}}$, are the measure of anchors.
$\pi$ is the transport plan, $\Pi(\cdot,\cdot)$ represent set of total transport permutation combination.
$c_{s}$ and $c_{t}$ are specific intra-distance metrics defined on the radial structures of the source and the target, respectively.
To incorporate the discriminative information, we recall classifier formulation $p(y=k|\z_{i})\propto\exp(\|W_{k}\|\|\z_{i}\|\cos(W_{k},\z_{i})+b)$ suggests angular and norm value is critical for vectors discrimination, we combine the both information in intra-distance function for $c_{s},c_{t}$, and define them with same formulation:
\begin{equation}
    c(\v_{i},\v_{j})=[1-\frac{\langle\v_{i},\v_{j}\rangle}{\|\v_{i}\|\|\v_{j}\|}] +\lambda_{\rm dist}\frac{1}{2} \|\v_{i}-\v_{j}\|^{2}_{2},
    \label{eq:intra-distance}
\end{equation}
with $\lambda_{\rm dist}$ weight parameter to tradeoff angular difference loss between the structures.
The first term calculates the cosine distances between the corresponding pairs of displacement vectors as angular difference. 
The second term captures the length difference by calculating the $\ell_2$ distances between the corresponding pairs of displacement vectors.
Furthermore, the transport plan $\pi$ can be fixed due to one-to-one correspondences of discriminative vectors $\mathcal{V}^{s}$ and $\mathcal{V}^{t}$ from source and target are known, i.e. we force $\pi_{i,j}=0$ when $i\neq j$ which gives:
\begin{equation*}
    GW(\mathcal{V}^{s},\mathcal{V}^{t})=\sum_{ij}|c(\v^{s}_{i},\v^{s}_{j})-c(\v^{t}_{i},\v^{t}_{j})|^{2},
\end{equation*}
where the GW distance with a fixed transport plan implies that a certain property of GW are lost, that is rotational invariance.
However, a loss of rotational difference yields a metric of shape difference that is useful for optimizing the Stiefel layer.
This completed the distance metric $d(\cdot, \cdot)$ in Eq.\eqref{eq:rotation_distance}.
Which has the advantage of providing a more efficient formula for the loss of local alignment and we defined it as $\phi(\mathcal{V}^{s},\mathcal{V}^{t})$.
Where the $\phi(\mathcal{V}^{s},\mathcal{V}^{t})$ can be expressed by the expectation of pairs of elements difference across two domains:
\begin{equation}
  \label{eq:strucutral_similarity}
  \phi(\mathcal{V}^{s},\mathcal{V}^{t}) = \mathbb{E}_{(\v^{s}_{k}, \v^{t}_{k})\in(\mathcal{V}^{s},\mathcal{V}^{t})} [c(\v^{s}_{k},\v^{t}_{k})].
\end{equation}
The simplified objective Eq.\eqref{eq:strucutral_similarity} gradually forces corresponding vectors being the same length and pointing to the same direction, thus ensuring the local structure alignment.

With synchronously minimizing the discrepancy among domains based on the radial structures by isometric transformation and structure refinement, data distributions of the source and target will move towards to each other and finally present the identical radial structure.
In this way, the posterior probability expectations of each category in the source and the target domains can also be consistent.

\subsection{Radial Structure Enhancement}

We further improve the learning of radial structures from the following two aspects;
1) First one is structure faithfulness requirement, which encourages samples to enclose their corresponding local anchors.
2) Second one is semantic meaningfulness requirement, extracted radial structure should be informative for the semantic information of data distribution, \emph{i.e.}, the consensus between geometrical assignment labels and classifier labels.

\subsubsection{Enclose features to local anchors}
According to the structure faithfulness requirement, features are expected to be located near desired anchors.
Since the distribution of features is unknown, we model the assignment of features to desired anchors by optimal transport plan \cite{villani2009optimal}.
In the case where the distances between features and anchors determine the transport cost, the optimal transport plan is the one which has the lowest total cost (also referred as optimal transport distance or Wasserstein distance) for moving features to corresponding anchors.
The optimal transport plan can also be viewed as an adaptive distribution model allowing different anchors correspond to different probability densities.
Then, to fairly push instances toward the desired local anchors, shared backbone network is learned to minimize optimal transport.
Further, for relaxing the objective and stablizing the end-to-end training we use entropic optimal transport \cite{cuturiSinkhornDistancesLightspeed2013a} distance defined by:
\begin{equation}
\begin{split}
    {\rm OT}^{\epsilon}_{\theta}&(\mathcal{X},\mathcal{N})=\\ &\min_{\pi\in\Pi(\mu,\mu_{a})}\sum_{i,j} d(G_{\theta}(\x_{i}),\a_{j})\pi_{i, j} + \epsilon {\rm KL}(\pi \|\mu\otimes\mu_{a}),    
\end{split}
\label{eq:local_optimal_transport}
\end{equation}
where $\mu=\sum^{n}_{i=1}\delta_{\x_{i}}$ the measure of data instances and $\mu_{a}=\sum^{k}_{j=1}\delta_{\a_{j}}$ the measure of anchors, $d(\cdot, \cdot)$ is euclidean distance metric, $\pi$ is the transport plan, $\Pi(\cdot,\cdot)$ represent set of total transport permutation combination, $\epsilon\geq 0$ is the regularization coefficient.
As a relevant metric for assigning samples to the best-fitted anchors, optimal transport distance can lead to a more reliable assignment than nearest neighbor search \cite{yang2021clustering}.
Therefore, by optimizing $G_{\theta}(\cdot)$ in minimizing $\mathcal{L}_{ot}={\rm OT}^{\epsilon}_{\theta}(\mathcal{X}^{s},\mathcal{N}^{s})+{\rm OT}^{\epsilon}_{\theta}(\mathcal{X}^{t},\mathcal{N}^{t})$ in both source and target domain independently, the extracted features in both domains are more compactly arranged around their radial structures, the structure faithfulness requirement can be indirectly achieved.

\subsubsection{Consensus regularization} 
For semantic meaningfulness requirement, we regard that instances assigned to the same local anchor have the same label, and for each instance, the label assigned by the classifier must match the label assigned by the radial structure.
Hence, to train a network meets semantic meaningfulness requirements, a consensus regularization is designed to force the labels assigned by the classifier match the labels assigned by the radial structure,
\begin{equation}
    \mathcal{R}_{\varphi}(\mathbf{Q},\mathbf{P}) = {\rm KL}(\mathbf{Q}||\mathbf{P}) + H(\mathbf{P}),
    \label{eq:consensus_regularization}
\end{equation}
where regularization is performed at classifier parameter $\varphi$, ${\rm KL}(\cdot||\cdot)$ is Kullback-Leibler divergence, $H(\cdot)$ is entropy that balances the discriminability negative effects in this regularization, $\mathbf{Q}=\{q_{i,k}\}$ is soft-assignments given by the transport plan $\pi$, $\mathbf{P}=\{(p_{i,1},\dots,p_{i,k})\}\in[0, 1]^{K\times N}$ indicates the posteriors given by classifier.
Consensus between data distribution structures and classifications can be improved by minimizing terms of regularization $\mathcal{L}_{R}=\mathcal{R}_{\varphi}(\mathbf{Q}^{s},\mathbf{P}^{s})+\mathcal{R}_{\varphi}(\mathbf{Q}^{t},\mathbf{P}^{t})$.

Intuitively, the objective based on Eq.\eqref{eq:local_optimal_transport} and Eq.\eqref{eq:consensus_regularization} gradually enhances the representative and discriminative of radial structures in each domain through minimizing optimal transport distance from samples to local anchors and consensus regularization between radial structure assignment and classification.

\subsection{Optimization}

The optimization is conducted in two steps, \textit{i.e.}, radial structures extraction and alignment.

\paragraph{Radial structure update}
The ideal implementation of calculating local anchors in Eq.\eqref{eq:local_anchors_calculation} requires iterating over the entire dataset, which is computationally expensive. By employing an appropriate exponential moving average update strategy, we can easily perform end-to-end training:
\begin{equation}
  \mathbf{a}_{k}=  \eta\frac{1}{M_{k}}\sum^{B}_{i=1}\z_{i}\mathbb{I}[y_{i}=k]+ (1-\eta) \mathbf{a}_{k}',
  \label{eq:ema}
\end{equation}
where $B$ indicates the batch size and $M_{k}=\sum^{B}_{i}\mathbb{I}[y_{i}=k]$ is a normalization constant, $\mathbf{a}_{k}'$ indicates the last updated anchors and $\mathbf{a}_{k}$ indicates new anchors computed in current iteration.

\begin{algorithm}[ht]
  \SetAlgoLined
  \KwData{Labeled source $\mathcal{D}^{s}$, Unlabeled target $\mathcal{D}^{t}$}
  \KwResult{$\theta$, $\varphi$, $\Delta$}
  Initialization: $\theta\leftarrow\theta_{0},\varphi\leftarrow\varphi_{0},\Delta\leftarrow\mathbf{I}$\;
  \While {Not Converge}{
    Sample $\{(\mathcal{X}^{s},\mathcal{Y}^{s})\}$ and $\{\mathcal{X}^{t}\}$ from $\mathcal{D}^{s}$ and $\mathcal{D}^{t}$\;
    $(\hat{\mathbf{P}}^{s},\hat{\mathbf{P}}^{t})\leftarrow (f_{\varphi}(G(\mathcal{X}^{s}),f_{\varphi}(S(G(\mathcal{X}^{t})))))$\;
    Update radial structures $\mathcal{G}^{s},\mathcal{G}^{t}$ according to Eq.\eqref{eq:local_anchors_calculation},Eq.\eqref{eq:domain_anchor}\;
    Calculate source classification loss  $\mathcal{L}_{ce}(\hat{\mathbf{P}}^{s},\mathcal{Y}^{s})$ \;
    Calculate alignment loss $\mathcal{L}_{\rm global},\phi(\mathcal{V}^{s},\mathcal{V}^{t})$ according to Eq.\eqref{eq:translation_distance}, Eq.\eqref{eq:strucutral_similarity} \;
    Calculate OT distance $\mathcal{L}_{ot}$ by Eq.\eqref{eq:local_optimal_transport}\;
    Calculate prediction discrepancy $\mathcal{L}_{R}$ between classifier and radial structure in Eq.\eqref{eq:consensus_regularization}\;
    // Update parameters according to gradients\;
    $\Delta\stackrel{+}\leftarrow -\nabla_{\Delta}\lambda_{\phi}\phi(\mathcal{V}^{s},\mathcal{V}^{t})$\;
    $\varphi\stackrel{+}\leftarrow  -\nabla_{\varphi}(\mathcal{L}_{ce}+\lambda_{R}\mathcal{L}_{R})$\;
    $\theta\stackrel{+}\leftarrow -\nabla_{\theta}(\mathcal{L}_{ce}+\lambda_{\rm ot}\mathcal{L}_{\rm ot}+\lambda_{\rm T}\mathcal{L}_{\rm global}+\lambda_{\phi}\phi(\mathcal{V}^{s},\mathcal{V}^{t}))$;}
  \Return {$\theta,\varphi,\Delta$}
  \caption{DRDA Training}
\end{algorithm}

\paragraph{Network update}

Recall objective of Eq.\eqref{eq:local_optimal_transport} and Eq.\eqref{eq:consensus_regularization}, a critical insights on behind successful optimization is similar to Expectation–Maximization (EM) algorithm.
To optimize optimal transport distance from samples to local anchors, we fixed local anchors and update $\theta$ according to Eq.\eqref{eq:local_optimal_transport}, then update local anchors make use of updated $\theta$ next iteration according to Eq.\eqref{eq:local_anchors_calculation}.
To optimize the consensus between geometrical assignments and classifier assignments, we fixed $\mathbf{Q}$ and only update classifier $\varphi$ according to Eq.\eqref{eq:consensus_regularization} with insights that classifier shall make trade off to respect intrinsic data distribution.
Finally, alternative network update approach can be easily implemented by \textit{stop gradient} tricks, then the overall objective respectively:
\begin{equation}
  \begin{split}
  \min_{\theta,\varphi,\Delta}~&\mathcal{L}_{ce}+\lambda_{T}\mathcal{L}_{\rm global} + \lambda_{\phi}\phi(\mathcal{V}^{s},\mathcal{V}^{t})\\
  &+\lambda_{ot}[OT^{\epsilon}_{\theta}(\mathcal{X}^{s},{\rm SG}[\mathcal{N}^{s}])+OT^{\epsilon}_{\theta}(\mathcal{X}^{t},{\rm SG}[\mathcal{N}^{t}])]\\
  &+\lambda_{R}[\mathcal{R}_{\varphi}({\rm SG}[\mathbf{Q}^{s}],\mathbf{P}^{s}) + \mathcal{R}_{\varphi}({\rm SG}[\mathbf{Q}^{t}],\mathbf{P}^{t})],
  \end{split}
  \label{eq:overall_objective}
\end{equation}
where ${\rm SG}[\cdot]$ indicates the stop-gradient operation. This operation prevents parameters from being updated by the gradients. In the light of alternative network update approach, stop-gradient operation is critical for preventing degeneration of the structure during learning.
Specifically, in Eq.\eqref{eq:overall_objective}, first term $\mathcal{L}_{ce}$ is classification error; the second term $\mathcal{L}_{\rm global}$ and third term $\phi(\mathcal{V}^{s},\mathcal{V}^{t})$ jointly perform isometric transformation and structure refinement for aligning feature distributions of different domains;
the rest terms enhance the representativity and discriminability of radial structures.
To balance the scale of terms in overall objective, the global loss (i.e. global translation distance) is scaled by $\lambda_{T}$, intra-structures difference is scaled by $\lambda_{\phi}$, OT distance is scaled by $\lambda_{ot}$ and consensus regularization is scaled by $\lambda_{R}$.
Notice, based upon differences Eq.\eqref{eq:strucutral_similarity} in the radial structures, the global rotation transformation distance minimization is implicitly optimized with respect to Stiefel layer parameters.

\begin{table*}[ht]
  \centering
  \caption{Accuracy ($\%$) on Office-31 for unsupervised domain adaptation (ResNet-50)}
  \label{tab:office-31}
  \resizebox{\textwidth}{!}{%
  \begin{tabular}{@{}cccccccc@{}}
  \toprule
  Method        & A $\rightarrow$ W      & D $\rightarrow$ W      & W $\rightarrow$ D      & A $\rightarrow$ D      & D $\rightarrow$ A      & W $\rightarrow$ A      & Average \\ \midrule
  ResNet-50      & 68.4 $\pm$ 0.2 & 96.7 $\pm$ 0.1 & 99.3 $\pm$ 0.1 & 68.9   $\pm$   0.2 & 62.5   $\pm$   0.3 & 60.7   $\pm$   0.3 & 76.1 \\
  RevGrad \cite{ganinUnsupervisedDomainAdaptation2015} & 82.0 $\pm$ 0.4 & 96.9 $\pm$ 0.2 & 99.1 $\pm$ 0.1 & 79.7 $\pm$ 0.4     & 68.2 $\pm$ 0.4     & 67.4 $\pm$ 0.5     & 82.2 \\
  DAN \cite{longLearningTransferableFeatures2015}  & 80.5 $\pm$ 0.4 & 97.1 $\pm$ 0.2 & 99.6 $\pm$ 0.1 & 78.6 $\pm$ 0.2 & 63.6 $\pm$ 0.3 & 62.8 $\pm$ 0.2 & 80.4    \\
  JAN \cite{longDeepTransferLearning2017}  & 85.4 $\pm$ 0.3 & 97.4 $\pm$ 0.2 & 99.8 $\pm$ 0.2 & 84.7 $\pm$ 0.3 & 68.6 $\pm$ 0.3 & 70.0 $\pm$ 0.4 & 84.3    \\
  MADA \cite{peiMultiAdversarialDomainAdaptation2018} & 90.0 $\pm$ 0.2 & 97.4 $\pm$ 0.1 & 99.6 $\pm$ 0.1 & 87.8 $\pm$ 0.2 & 70.3 $\pm$ 0.3 & 66.4 $\pm$ 0.3 & 85.2    \\
  CDAN+E*\cite{longConditionalAdversarialDomain2018a} & 94.1 $\pm$ 0.1 & 98.6 $\pm$ 0.1 & 100.0 $\pm$ .0 & 92.9 $\pm$ 0.2 & 71.0 $\pm$ 0.3 & 69.3 $\pm$ 0.3 & 87.7 \\
  ALDA\cite{chenAdversarialLearnedLossDomain2020} & 95.6 $\pm$ 0.5 & 97.7 $\pm$ 0.5 & 100.0 $\pm$ .0 & 94.0 $\pm$ 0.4 & 72.2 $\pm$ 0.4 & 72.5 $\pm$ 0.2 & 88.7 \\
  \hline
  DRDA (w/o Angular) & 92.6 $\pm$ 0.5 & 98.3 $\pm$ 0.2 & 100.0 $\pm$ .0 & 91.9 $\pm$ 0.5 & 71.0 $\pm$ 0.3 & 70.6 $\pm$ 0.1 & 87.4 \\
  DRDA (w/o Stiefel) & 92.3 $\pm$ 0.5 & 98.7 $\pm$ 0.1 & 100.0 $\pm$ .0 & 92.1 $\pm$ 0.5 & 74.7 $\pm$ 0.2 & 75.3 $\pm$ 0.2 & 88.8 \\
  DRDA (w/o $\mathcal{R}_{\varphi}$) & 94.9 $\pm$ 0.3 & 98.2 $\pm$ 0.1 & 100.0 $\pm$ .0 & 93.8 $\pm$ 0.3 & 74.2 $\pm$ 0.1 & 75.8 $\pm$ 0.1 & 89.4\\
  DRDA (w/o ${\rm OT}^{\epsilon}_{\theta}$) & 94.8 $\pm$ 0.2 & 98.0 $\pm$ 0.1 & 100.0 $\pm$ .0 & 94.0 $\pm$ 0.2 & 74.8 $\pm$ 0.1 & 75.4 $\pm$ 0.1 & 89.5 \\
  DRDA & \textbf{95.8} $\pm$ 0.4 & \textbf{98.8} $\pm$ 0.4 & \textbf{100.0} $\pm$ .0 & \textbf{94.5} $\pm$ 0.3 & \textbf{75.6} $\pm$ 0.2 & \textbf{76.6} $\pm$ 0.4 & \textbf{90.2} \\
  \bottomrule
  \end{tabular}
  }
\end{table*}

\begin{table*}[ht]
  \caption{Accuracy ($\%$) on Office-Home for unsupervised domain adaptation (ResNet-50)}
  \label{tab:office-home}
  \resizebox{\textwidth}{!}{%
  \begin{tabular}{@{}cccccccccccccc@{}}
  \toprule
  Method & Ar$\rightarrow$Cl         & Ar$\rightarrow$Pr         & Ar$\rightarrow$Rw & Cl$\rightarrow$Ar         & Cl$\rightarrow$Pr       & Cl$\rightarrow$Rw & Pr$\rightarrow$Ar         & Pr$\rightarrow$Cl         & Pr$\rightarrow$Rw & Rw$\rightarrow$Ar         & Rw$\rightarrow$Cl         & Rw$\rightarrow$Pr         & Average      \\ \midrule
  ResNet-50 & 34.9 & 50.0 & 58.0 & 37.4 & 41.9 & 46.2 & 38.5 & 31.2 & 60.4 & 53.9 & 41.2 & 59.9 & 46.1 \\
  DANN \cite{ganinDomainAdversarialTrainingNeural2016} & 45.6 & 59.3 & 70.1 & 47.0 & 58.5 & 60.9 & 46.1 & 43.7 & 68.5 & 63.2 & 51.8 & 76.8 & 57.6 \\
  JAN \cite{longDeepTransferLearning2017}    & 45.9 & 61.2 & 68.9 & 50.4 & 59.7 & 61.0 & 45.8 & 43.4 & 70.3 & 63.9 & 52.4 & 76.8 & 58.3 \\
  CDAN+E \cite{longConditionalAdversarialDomain2018a} & 50.7 & 70.6 & 76.0 & 57.6 & 70.0 & 70.0 & 57.4 & 50.9 & 77.3 & 70.9 & 56.7 & 81.6 & 65.8 \\
  ALDA \cite{chenAdversarialLearnedLossDomain2020}   & 53.7          & 70.1          & 76.4  & 60.2          & 72.6        & 71.5  & 56.8          & 51.9          & 77.1  & 70.2          & 56.3          & 82.1          & 66.6     \\
  MDD \cite{zhangBridgingTheoryAlgorithm2019}  & 54.9 & 73.7 & 77.8 & 60.0 & 71.4 & 71.8 & 61.2 & 53.6 & 78.1 & 72.5 & 60.2 & 82.3 & 68.1 \\
  \hline
  DRDA (w/o Angular) &
  54.3 &
  70.3 &
  74.8 &
  60.7 &
  69.2 &
  69.8 &
  59.1 &
  52.8 &
  76.4 &
  70.9 &
  58.3 &
  82.0 &
  66.5 \\
  DRDA (w/o Stiefel) &
  57.4 &
  74.5 &
  79.3 &
  64.8 &
  75.6 &
  \textbf{74.0} &
  62.9 &
  56.2 &
  79.7 &
  72.0 &
  62.9 &
  84.1 &
  70.4 \\
  DRDA (w/o $\mathcal{R}_{\varphi}$) &
  57.3 &
  74.3 &
  80.4 &
  64.7 &
  74.3 &
  73.0 &
  64.9 &
  55.8 &
  79.7 &
  74.5 &
  63.0 &
  84.3 &
  70.5\\
  DRDA (w/o ${\rm OT}^{\epsilon}_{\theta})$ &
  57.0 &
  73.9 &
  80.2 &
  64.1 &
  73.8 &
  73.1 &
  64.4 &
  56.1 &
  78.9 &
  73.4 &
  62.8 &
  84.1 &
  70.1 \\
  DRDA &
  \textbf{58.2} &
  \textbf{74.2} &
  \textbf{81.2} &
  \textbf{65.6} &
  \textbf{75.1} &
  73.3 &
  \textbf{65.8} &
  \textbf{57.1} &
  \textbf{80.4} &
  \textbf{75.6} &
  \textbf{63.2} &
  \textbf{85.1} &
  \textbf{71.2} \\ \bottomrule
  \end{tabular}%
  }
\end{table*}
\section{Experiments}

We compare the proposed method with several state-of-art methods on three types of UDA tasks, including single UDA, Domain-Agnostic UDA and Multi-Source UDA. The experimental results show that our method outperforms the other methods in terms of the average classification accuracy. In addition, we present a series of visualization results and ablation studies to demonstrate the insights of our method and the effectiveness of each  component in our model.

\subsection{Experimental Setup}
\subsubsection{Office-31} \cite{saenkoAdaptingVisualCategory2010}
is a widely used dataset for visual domain adaptation, which consists of 31 categories count up to 4,652 images from three distinct domains: 2,817 Amazon\textbf{(A)} images, 795 Webcam\textbf{(W)} images, and 498 DSLR\textbf{(D)} images. We evaluate methods upon all $6$ in pairs of transfer tasks.
\subsubsection{Office-Home} \cite{Venkateswara_2017_CVPR}
is a better organized dataset and more difficult dataset compared to Office-31, which consists of 65 categories count up to 15,500 images in office and home setting, formed with four extremely dissimilar domains: Artistic images \textbf{(Ar)}, Clipart images \textbf{(Cl)}, Product images \textbf{(Pr)}, and Real-World images \textbf{(Rw)}.
\subsubsection{Office-Caltech10} \cite{gongGeodesicFlowKernel2012}
is collected from Office31 and Caltech formed with four domains: \textbf{A} (Amazon), \textbf{C} (Caltech), \textbf{W} (Webcam), and \textbf{D} (DSLR). It consists of 10 object categories, each domain includes 958, 295, 157, and 1,123 images, respectively.

We compared the proposed~\textbf{(DRDA)} with state-of-the-art domain adaptation methods: Domain Adversarial Neural Network \textbf{(DANN)} \cite{ganinDomainAdversarialTrainingNeural2016}, Joint Adaptation Network \textbf{(JAN)} \cite{longDeepTransferLearning2017}, Conditional Domain Adversarial Network with Entropy \textbf{(CDAN+E)} \cite{longConditionalAdversarialDomain2018a}, Adversarial-Learned Loss for Domain Adaptation \textbf{(ALDA)} \cite{chenAdversarialLearnedLossDomain2020} and Margin Disparity Discrepancy \textbf{(MDD)} \cite{zhangBridgingTheoryAlgorithm2019}.
For \textbf{multi-source domain adaptation} we compared our model with state-of-the-art domain adaptation methods: Deep Alignment Network \textbf{(DAN)} \cite{longUnsupervisedDomainAdaptation2016}, Domain Adversarial Neural Network \textbf{(DANN)} \cite{ganinDomainAdversarialTrainingNeural2016}, Manifold Embedded Distribution Alignment \textbf{(MEDA)} \cite{wangVisualDomainAdaptation2018}, Maximum Classifier Discrepancy \textbf{(MCD)} \cite{saitoMaximumClassifierDiscrepancy2018} and Moment Matching for Multi-Source Domain Adaptation \textbf{(M$^{3}$SDA)} \cite{pengMomentMatchingMultiSource2019}.
For \textbf{domain agnostic domain adaptation}, we compared our model with state-of-the-art methods: Self-Ensembling \textbf{(SE)} \cite{xuSelfEnsemblingAttentionNetworks2019}, Maximum Classifier Discrepancy \textbf{(MCD)} \cite{saitoMaximumClassifierDiscrepancy2018}, Domain Adversarial Neural Network \textbf{(DANN)} \cite{ganinDomainAdversarialTrainingNeural2016} and Deep Adversarial Disentangled Autoencoder \textbf{(DADA)} \cite{pengDomainAgnosticLearning2019}.

We follow the standard protocols of unsupervised domain adaptation. We use all labeled source samples and unlabeled target samples and compare the average classification accuracy based on three experiments. The overall architecture
consists of a backbone, \textbf{ResNet-50}, a bottleneck layer with $256$ units and a full-connected layer. 
The Stiefel layer is a simple full-connected layer whose parameters are manipulated on \textit{Stiefel Manifold} implemented with \emph{geoopt}\cite{kochurovGeooptRiemannianOptimization2020}.
And this Stiefel layer is used for processing target features only.
We implement our method in \textbf{Pytorch}. 
We finetune from ImageNet pre-trained models as the feature extract backbone.
We essentially tune the hyper-parameters in Eq.\eqref{eq:overall_objective}, $\lambda_{T}\sim 200,\lambda_{\phi}\sim 0.6,\lambda_{ot}\sim 0.0005,\lambda_{R}\sim 1$, they control the scaling of each loss term in overall objective.
Both backbone layers and task-specific layers are trained through back-propagation using Stochastic Gradient Descent (SGD).
The Stiefel layer is optimized using Riemannian SGD\cite{kochurovGeooptRiemannianOptimization2020}.
The backbone layers is finetuned based on pre-trained ResNet-50 on ImageNet, while the task-specific layers are trained from scratch whose learning rate is 10 times that of backbone layers.

We use mini-batch stochastic gradient descent as the optimizer and apply momentum of 0.9 and learning rate schedule rule \cite{ganinDomainAdversarialTrainingNeural2016} with $\eta_{p}=\eta_{0}(1+\gamma p)^{-\beta}$, where $p$ is within the range of $[0, 1]$, and $\eta_{0}=0.01$, $\gamma=10$, $\beta=0.75$.
We conduct the grid hyper-parameter selection base on loss curve fitting to obtain optimal weighted combination of objective in the experiments. To reduce noise influence and stable optimization convergence, we adopt progressive domain transfer weights with factor $\lambda_{p}=\frac{2}{1+\exp(-\alpha p)}-1$ increasing from 0 to 1 where $\alpha=10$ as a default setting. Especially, for Office31 dataset, due to the small number of samples in DSLR and Webcam domain, we add temperature factor $t=0.85$ to adjust the convergence speed of a cross-entropy loss.

For the experiments on Multiple source unsupervised domain adaptation using Office-Caltech and OfficeHome, we sample instances uniformly from the combined source domains as source inputs.
For the experiments on Domain agnostic unsupervised domain adaptation using Office-Caltech we sample instances uniformly from the combined target domains as target inputs.
For the experiments on domain-generalized unsupervised domain adaptation, we train models similar to the setting of one-to-one UDA, and demonstrate the validation accuracies in domains which were not the sources or targets.

\subsection{Comparison with the State-of-The-Art Methods}

\subsubsection{Single source to single target UDA}

To testify the effectiveness of the proposed DRDA, we first compare our method with state-of-art single domain UDA tasks.
For a fair comparison, we report previous domain adaptation methods whose results are based on ResNet-50 and test in same validation setting.

The results on Office-31 are reported in Table \ref{tab:office-31}.
In most transfer sub-tasks, DRDA attains the highest classification accuracy and improves the average accuracy over state-of-the-art methods.
Our method works particularly well for small-to-large transfer tasks, such as \textbf{D}$\rightarrow$\textbf{A},\textbf{W}$\rightarrow$\textbf{A}.
Even though the sample size in the source domain is small, the proposed discriminative radial structure is sufficiently representative and discriminative to serve as a guide to domain alignment and more robust to noise.

The results on Office-Home are reported in Table \ref{tab:office-home}.
The proposed DRDA achieves the best accuracy in all transfer tasks and improves the average accuracy over the state-of-the-art methods by $3\%$. Compared to Office-31, Office-Home is more challenging because it has more categories and greater discrepancies between domains.
As the task becomes increasingly challenging, our approach outperforms our competitors by a greater margin.
As explained in the hypothesis, radial-like structures are beneficial for sketching and preserving discriminative structures, and this structure is well suited for domain alignment.
Therefore, in the case of more categories and greater discrepancies between domains, the radial-like structure shows greater superior performance in domain alignment than the other methods. In the context of domain alignment tasks, these results illustrate the importance of discrimination preservation and low-dimensional structures (i.e. the proposed radial-like structure).

\begin{table}[ht]
\centering
\caption{Accuracy ($\%$) on Office-Caltech for Multi-Source unsupervised domain adaptation (ResNet-50)}
\label{tab:multi_domain}
\resizebox{\linewidth}{!}{%
\begin{tabular}{@{}cccccc@{}}
\toprule
Method  & A,C,D$\rightarrow$W & A,C,W$\rightarrow$D & A,D,W$\rightarrow$C & C,D,W$\rightarrow$A & Avg  \\ \midrule
ResNet-50 & 97.1 & 99.2 & 89.4 & 94.7 & 95.6 \\
DANN \cite{ganinDomainAdversarialTrainingNeural2016} & 96.5 & 99.1 & 89.2 & 94.7 & 94.8 \\
MEDA \cite{wangVisualDomainAdaptation2018}  & 99.3    & 99.2    & 91.4    & 92.9    & 95.7 \\
MCD \cite{saitoMaximumClassifierDiscrepancy2018} & 99.5    & 99.1    & 91.5    & 92.1    & 95.6 \\
M$^3$SDA-$\beta$ \cite{pengMomentMatchingMultiSource2019} & 99.5    & 99.2    & 92.2    & 94.5    & 96.4 \\
\midrule
DRDA (w/o Angular) & 100.0 & 100.0 & 95.7 & 96.5 & 98.1 \\
DRDA (w/o Stiefel) & 100.0 & 100.0 & 95.7 & 96.8 & 98.1 \\
DRDA (w/o $\mathcal{R}_{\varphi}$) & 100.0 & 100.0 & 95.8 & 96.5 & 98.0 \\
DRDA (w/o ${\rm OT}^{\epsilon}_{\theta}$) & 100.0 & 100.0 & 96.0 & 96.8 & 98.2 \\
DRDA (ours)  &  \textbf{100.0}  &  \textbf{100.0}  & \textbf{96.4}  & \textbf{96.9}   & \textbf{98.3} \\ \bottomrule
\end{tabular}%
}
\end{table}
\subsubsection{Multi-source to single target UDA}
\textbf{multi-source unsupervised domain adaptation}
\cite{pengMomentMatchingMultiSource2019}, which transfers knowledge from multiple source domains to one unlabeled target domain. Compared to one-to-one unsupervised domain adaptation, this task is much more difficult as the source domain is a mixture of multiple domains.
In this task, we merge the multiple source domain into a single one.
The results on the Office-Caltech10 dataset are reported in Table~\ref{tab:multi_domain}.
According to the observation, the proposed DRDA surpasses state-of-the art methods even those developed for such tasks specifically. Recall the conception of radial-like structure, the key idea is sketching the discriminative structure of data distributions. In this respect, the more domains the algorithm uses as sources, the greater the generalization power the source structure has. Therefore, the proposed DRDA is naturally fit for UDA tasks involving multiple source domains.

\begin{table}[ht]
\centering
\caption{Accuracy ($\%$) on Office-Caltech for Domain-Agnostic unsupervised domain adaptation (ResNet-50)}
\label{tab:domain_agnostic}
\resizebox{\linewidth}{!}{%
\begin{tabular}{@{}cccccc@{}}
\toprule
Method & A$\rightarrow$ C,D,W & C$\rightarrow$ A,D,W & D$\rightarrow$ A,C,W & W$\rightarrow$ A,C,D & Avg \\ \midrule
ResNet-50 & 90.5$\pm$0.3 & 94.3$\pm$0.2 & 88.7$\pm$0.4 & 82.5$\pm$0.3 & 89   \\
SE \cite{xuSelfEnsemblingAttentionNetworks2019}  & 90.3$\pm$0.4 & 94.7$\pm$0.4 & 88.5$\pm$0.3 & 85.5$\pm$0.4 & 89.7 \\
MCD \cite{saitoMaximumClassifierDiscrepancy2018} & 91.7$\pm$0.4 & 95.3$\pm$0.3 & 89.5$\pm$0.2 & 84.3$\pm$0.2 & 90.2 \\
DANN \cite{ganinDomainAdversarialTrainingNeural2016} & 91.5$\pm$0.4 & 94.3$\pm$0.4 & 90.5$\pm$0.3 & 86.3$\pm$0.3 & 90.6 \\
DADA \cite{pengDomainAgnosticLearning2019} & 92.0$\pm$0.4 & 95.1$\pm$0.3 & 91.3$\pm$0.4 & 93.1$\pm$0.3 & 92.9 \\
\midrule
DRDA (w/o Angular) & 97.6$\pm$0.4 & 97.8$\pm$0.5 & 96.2$\pm$0.4 & 96.7$\pm$0.1 & 97.2 \\
DRDA (w/o Stiefel) & 97.7$\pm$0.3 & 97.1$\pm$0.7 & 96.8$\pm$0.1 & 97.0$\pm$0.2 & 97.2 \\
DRDA (w/o $\mathcal{R}_{\varphi}$) & 97.6$\pm$0.5 & 97.6$\pm$0.3 & 96.6$\pm$0.5 & 96.7$\pm$0.2 & 97.1 \\
DRDA (w/o $\rm{OT}^{\epsilon}_{\theta}$) & 97.8$\pm$0.5 & \textbf{98.0}$\pm$0.2 & \textbf{97.1}$\pm$0.5 & 96.8$\pm$0.3 & \textbf{97.5} \\
DRDA (ours)  & \textbf{98.1}$\pm$0.2 & 97.5$\pm$0.2 & 96.6$\pm$0.4 & \textbf{96.8}$\pm$0.2 & 97.3 \\ \bottomrule
\end{tabular}%
}
\end{table}

\begin{table*}[ht]
\caption{Accuracy (\%) on Office-Home for Domain Generalize unsupervised domain adaptation (ResNet-50)}
\label{tab:domain_generalization_officehome}
\resizebox{\textwidth}{!}{%
\begin{tabular}{@{}l|cc|cc|cc|cc|cc|cc|cc|cc|cc|cc|cc|cc|c@{}}
\toprule
Train &
  \multicolumn{2}{c|}{Ar→Cl} &
  \multicolumn{2}{c|}{Ar→Pr} &
  \multicolumn{2}{c|}{Ar→Rw} &
  \multicolumn{2}{c|}{Cl→Ar} &
  \multicolumn{2}{c|}{Cl→Pr} &
  \multicolumn{2}{c|}{Cl→Rw} &
  \multicolumn{2}{c|}{Pr→Ar} &
  \multicolumn{2}{c|}{Pr→Cl} &
  \multicolumn{2}{c|}{Pr→Rw} &
  \multicolumn{2}{c|}{Rw→Ar} &
  \multicolumn{2}{c|}{Rw→Cl} &
  \multicolumn{2}{c|}{Rw→Pr} &
  \multirow{2}{*}{Avg} \\ \cmidrule(r){1-25}
Test &
  Pr &
  Rw &
  Cl &
  Rw &
  Cl &
  Pr &
  Rw &
  Pr &
  Ar &
  Rw &
  Ar &
  Pr &
  Cl &
  Rw &
  Ar &
  Rw &
  Ar &
  Cl &
  Pr &
  Cl &
  Ar &
  Pr &
  Ar &
  Cl &
   \\ \midrule
  ResNet-50 & 50.0 & 58.0 & 34.9 & 58.0 & 34.9 & 50.0 & 46.2 & 41.9 & 37.4 & 41.9 & 37.4 & 41.9 & 31.2 & 60.4 & 38.5 & 60.4 & 38.5 & 31.2 & 59.9 & 41.2 & 53.9 & 59.9 & 53.9 & 41.2 & 46.1 \\
  DANN\cite{ganinDomainAdversarialTrainingNeural2016} & 60.0 & 68.7 & 37.8 & 70.0 & 42.3 & 66.3 & 63.0 & 59.1 & 49.4 & 64.3 & 54.8 & 63.2 & 37.4 & 71.9 & 48.0 & 68.0 & 58.6 & 39.7 & 75.1 & 41.2 & 59.2 & 73.3 & 63.1 & 43.4 & 57.4 \\
  MDD\cite{zhangBridgingTheoryAlgorithm2019} & 61.6 & 68.5 & 38.1 & 71.4 & 40.7 & 67.0 & 63.9 & 60.2 & 47.1 & 62.3 & 53.7 & 61.9 & 34.7 & 70.1 & 46.7 & 67.9 & 54.6 & 35.9 & 74.5 & 40.5 & 61.5 & 74.3 & 60.6 & 40.0 & 56.5 \\
  \midrule
DRDA &
  \textbf{65.6} &
  \textbf{74.1} &
  \textbf{50.6} &
  \textbf{75.8} &
  \textbf{50.3} &
  \textbf{72.1} &
  \textbf{71.6} &
  \textbf{68.9} &
  \textbf{60.1} &
  \textbf{71.4} &
  \textbf{61.7} &
  \textbf{68.9} &
  \textbf{48.2} &
  \textbf{76.9} &
  \textbf{59.7} &
  \textbf{75.7} &
  \textbf{64.4} &
  \textbf{51.2} &
  \textbf{80.7} &
  \textbf{52.9} &
  \textbf{70.8} &
  \textbf{79.3} &
  \textbf{72.0} &
  \textbf{52.2} &
  \textbf{65.6} \\ \bottomrule
\end{tabular}%
}
\end{table*}

\begin{figure*}[ht]
 \centering
 \includegraphics[width=\linewidth]{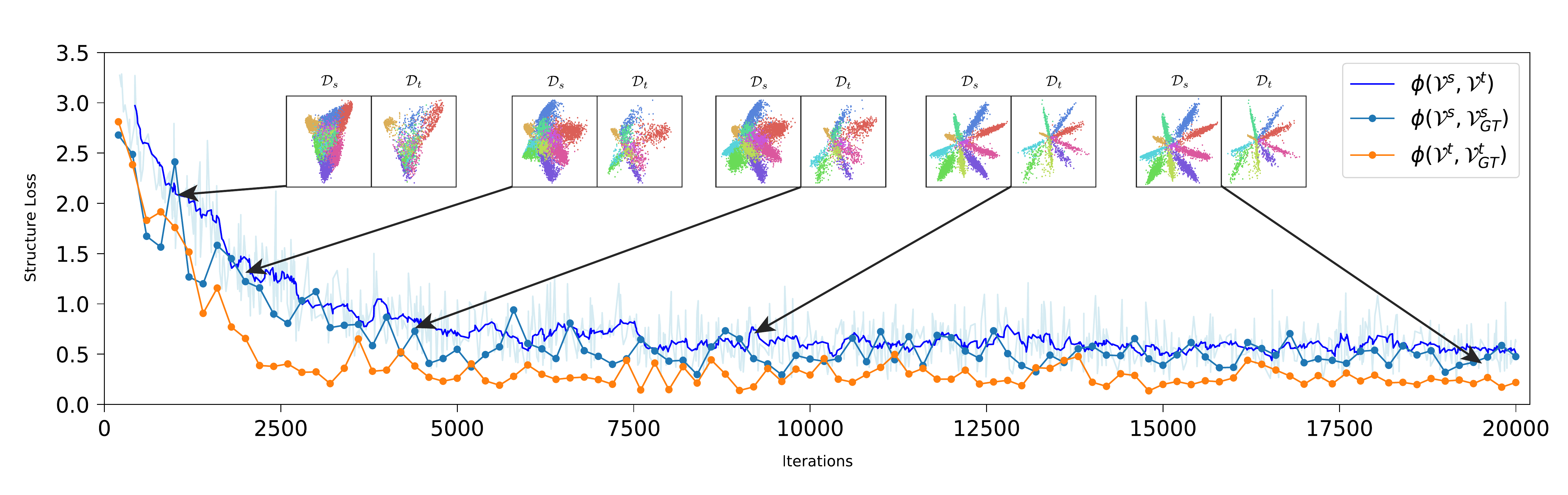}
 \caption{Training curve and latent features visualization at difference stages, the colors of points indicate instances colors. We can see a clear structure evolution progress while structure loss decreasing. Where $\phi(\mathcal{V}^{s},\mathcal{V}^{t}_{GT}$) is computed every validation, $\mathcal{V}^{s}_{GT}$ indicates the structure calculated with ground truth labels (resp $\phi(\mathcal{V}^{t},\mathcal{V}^{t}_{GT})$). $\phi(\mathcal{V}^{s},\mathcal{V}^{t})$ is passed through a median filter (original with light blue color) for visualization purpose, best view in color.}
 \label{fig:structure_evolution}
\end{figure*}

\subsubsection{Single source to agnostic multi-target UDA}
We also consider another type of unsupervised domain adaptation task:~\textbf{domain agnostic unsupervised domain adaptation} \cite{pengDomainAgnosticLearning2019}, which transfers knowledge from a labeled source domain to unlabeled data in one of multiple target domains.
In this task, we regard the mixture target domain as a single one.
The Table~\ref{tab:domain_agnostic} shows that our model gets a $97.3\%$ average accuracy and improves the other methods by $4.4\%$ in the domain agnostic unsupervised domain adaptation task. It appears that the radial-like structure is consistently effective at representing discriminative structures regardless of domain heterogeneity. Hence, the domain alignment can be well assured with the help of the radial structure.

\subsubsection{Domain generalize UDA}
A further extension to illustrate the utility of DRDA for knowledge abstraction is to extend it to domain-generalized UDA problems, that is, to train the model on one task and to test it on another domain that is different from the source and target domains.

The detailed results on OfficeHome were reported in Table~\ref{tab:domain_generalization_officehome}, where the first row indicates the standard single-to-single UDA  task that the models are trained for, and the second row indicates the test domain used only for evaluation.
These results indicate that our method is capable of generalizing to domain generalization tasks.
The DRDA method performed well in a number of subtasks. In many domains of the test, our method was superior to those classical UDA methods that directly optimize adaptation performance on those domains, even though our method did not incorporate this domain information for training.
These results presented here provide evidence for the effectiveness of radial-like structures in discriminative  feature modeling.
The in-depth explanation is that the alignment using discriminative radial structures forces the network to learn more meaningful features as a result of regularizing its optimization pathway.
As a consequence, when the trained model encounters instances from domains that have not previously been encountered, they can also be classified on the basis of their semantic features.

\subsection{Analysis}
In this subsection, we present various experiments that illustrate the intuition behind radial-like structures and demonstrate the effectiveness of our approach.
\begin{figure*}[ht]
  \centering
  \subfloat[Accuracy Convergence]{\includegraphics[width=0.24\linewidth]{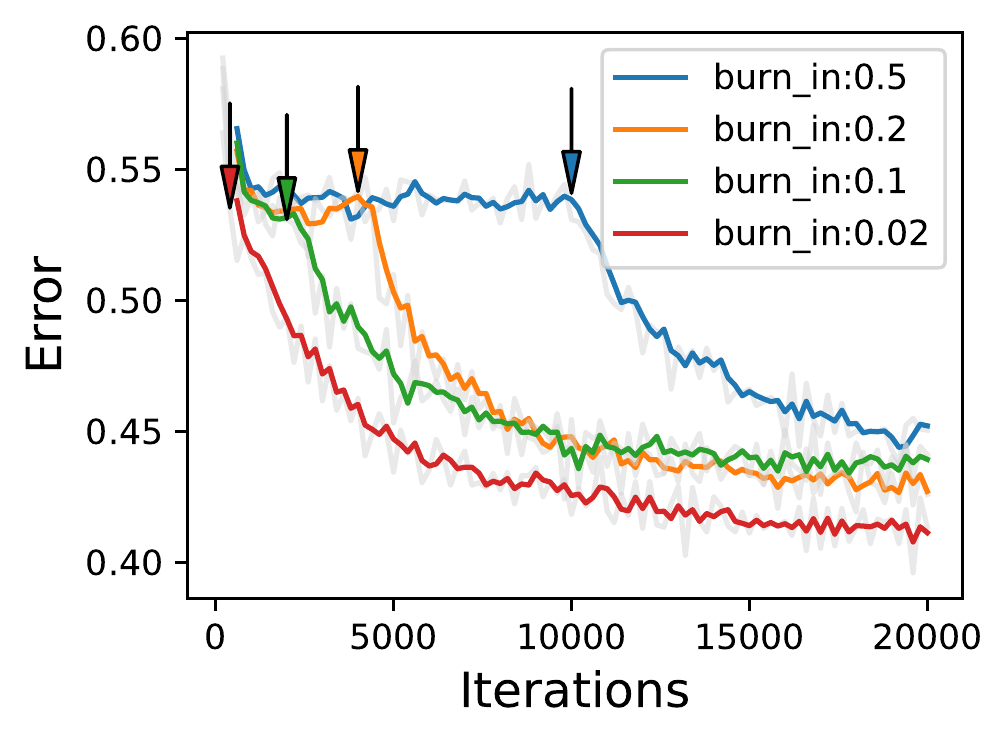}\label{fig:s2m_acc}}
  \subfloat[Structure Loss]{\includegraphics[width=0.24\linewidth]{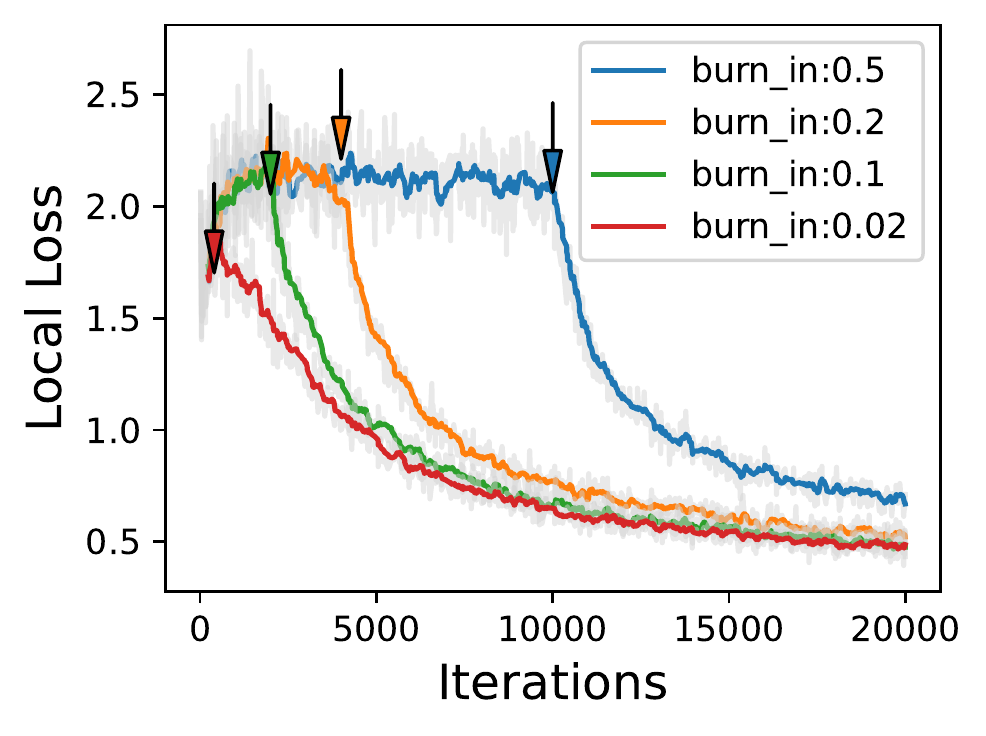}\label{fig:sub_structure_loss}}
  \subfloat[Wasserstein]{\includegraphics[width=0.24\linewidth]{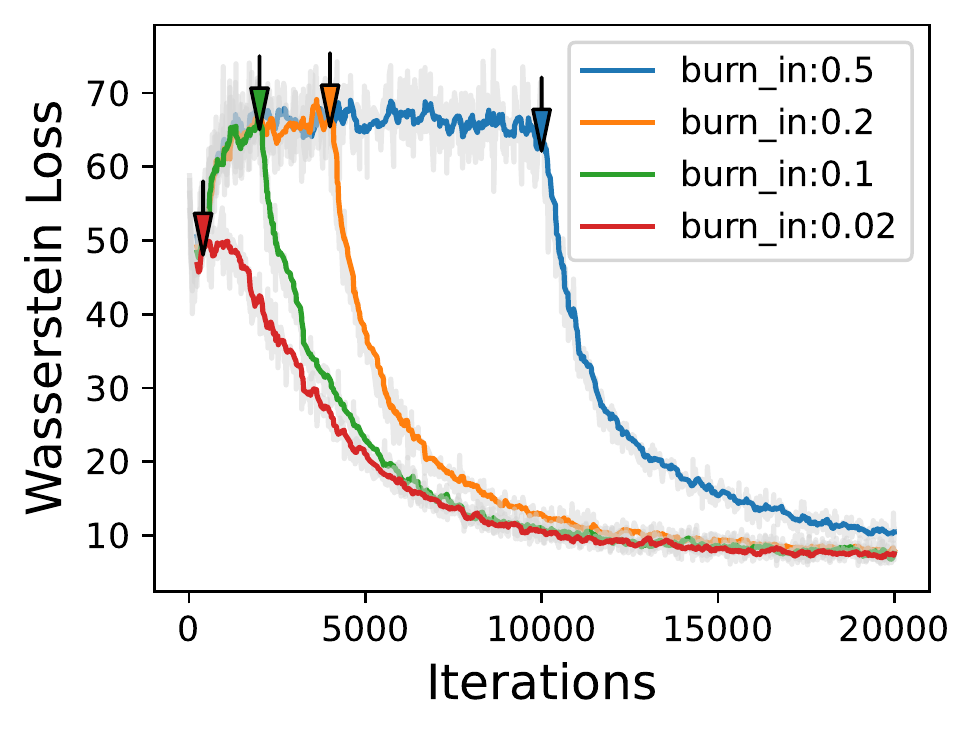}}
  \subfloat[KL Divergence]{\includegraphics[width=0.24\linewidth]{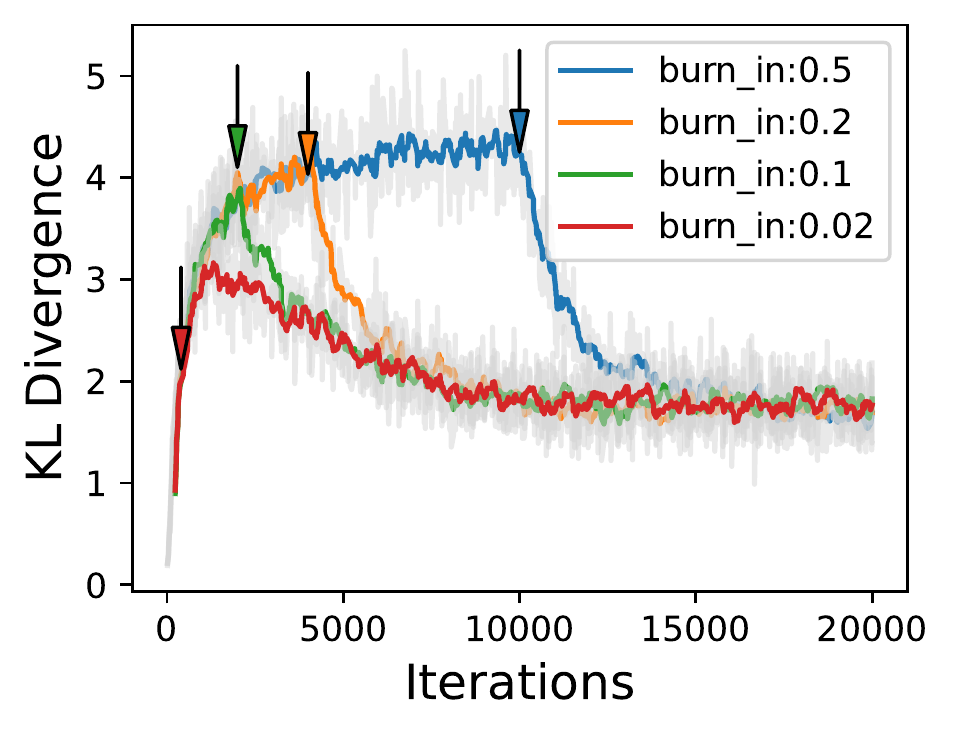}}
  \caption{The component loss along with the train iterations.
   We use `burn\_in' to indicate the percentage of training progress before adding the structure alignment operations. (a). validation accuracies of Clipart along with training iterations (b). $\phi(\mathcal{V}^{s},\mathcal{V}^{t})$ along with training iterations. (c). wasserstein distance between instances and local anchors in the target domain, along with training iterations. (d). The KL divergence between target geometrical label assignments and the classifier label assignments. (best view in color) }
  \label{fig:evolution_process}
\end{figure*}
\begin{figure*}
  \centering
  \subfloat[$\lambda_{T}$]{\includegraphics[width=0.24\linewidth]{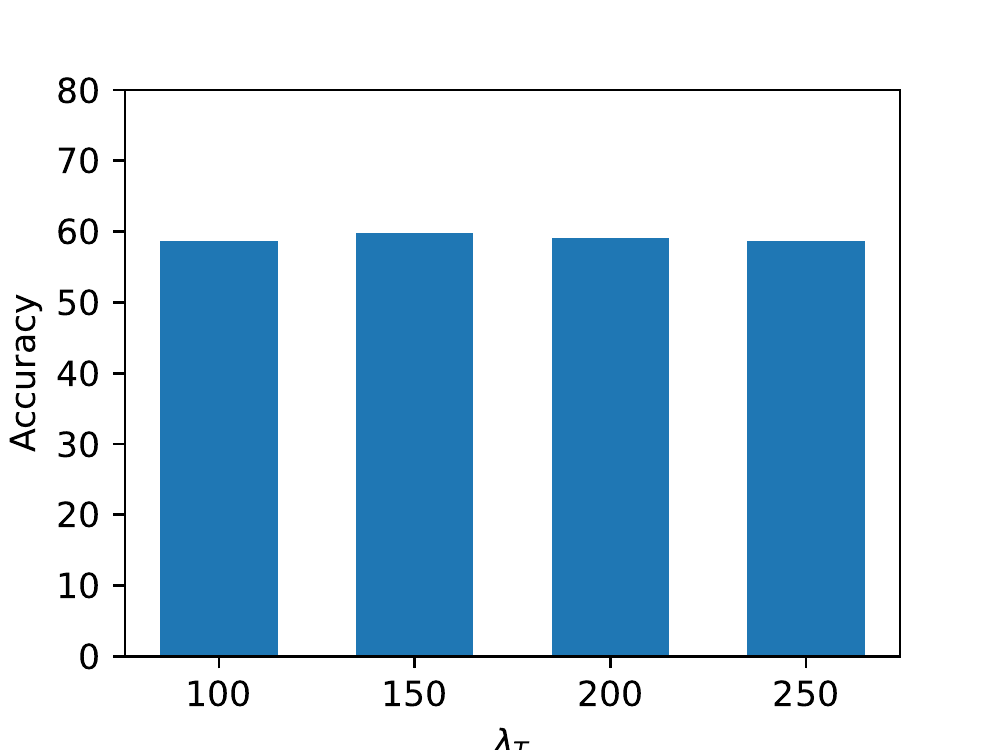}\label{fig:param_t}}
  \subfloat[$\lambda_{ot}$]{\includegraphics[width=0.24\linewidth]{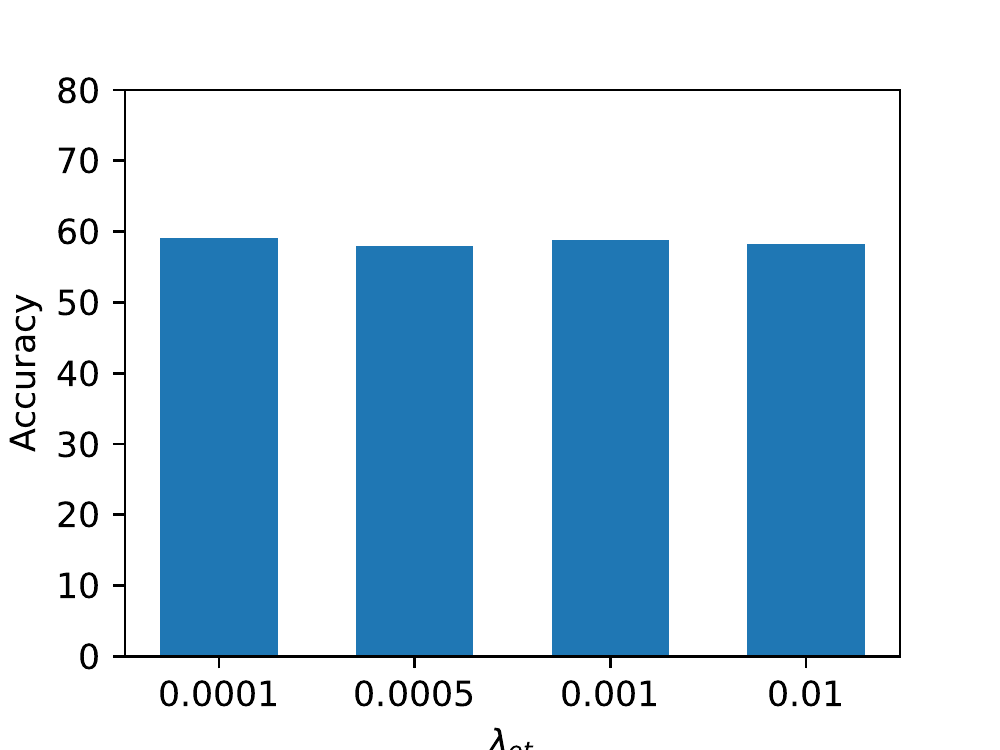}\label{fig:param_ot}}
  \subfloat[$\lambda_{\phi}$]{\includegraphics[width=0.24\linewidth]{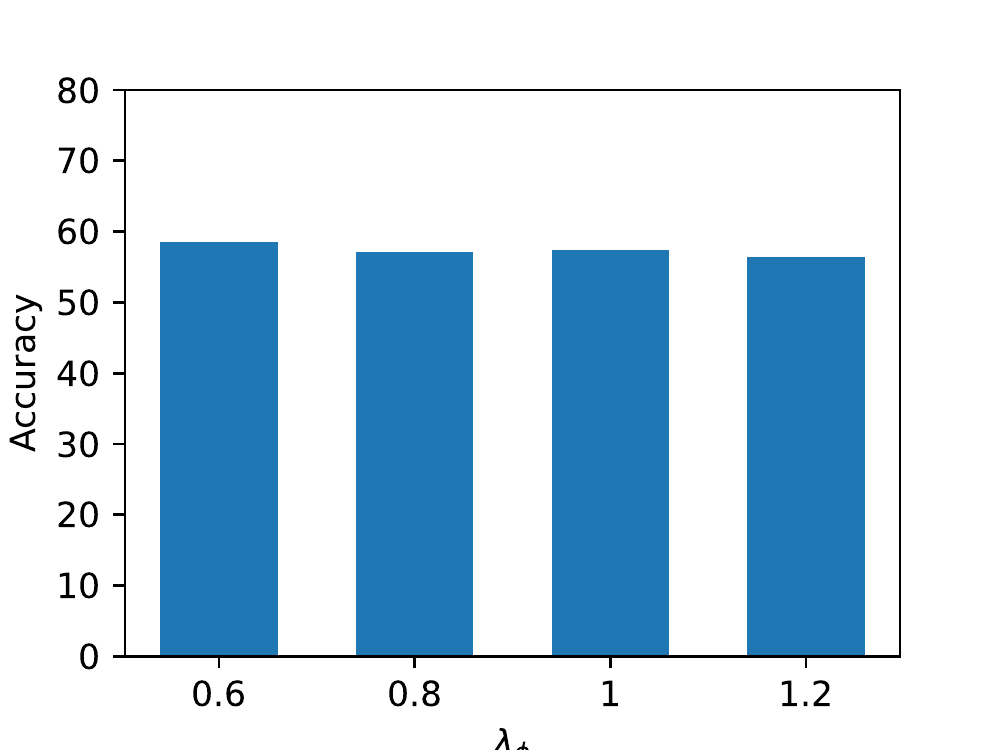}}\label{fig:param_phi}
  \subfloat[$\lambda_{R}$]{\includegraphics[width=0.24\linewidth]{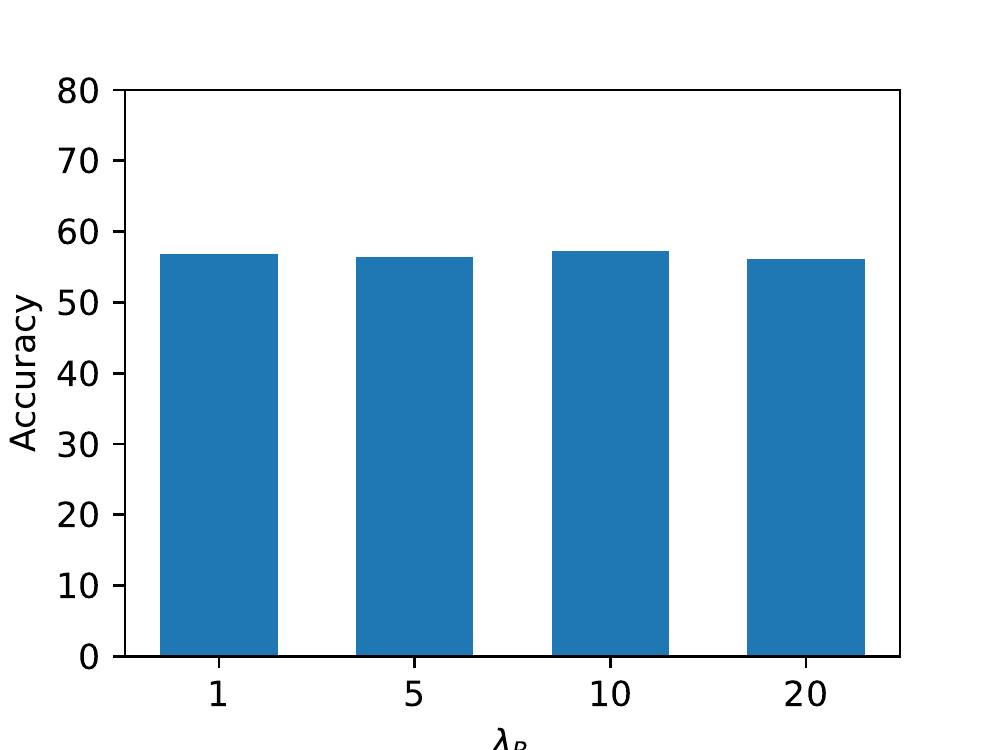}}\label{fig:param_R}
  \caption{The effect of different hyperparameters on the performance of OfficeHome dataset. Here we report subtask results on Art$\rightarrow$Clipart.}
  \label{fig:param_sensitivity}
\end{figure*}
\begin{figure}
    \centering
    \includegraphics[width=0.48\linewidth]{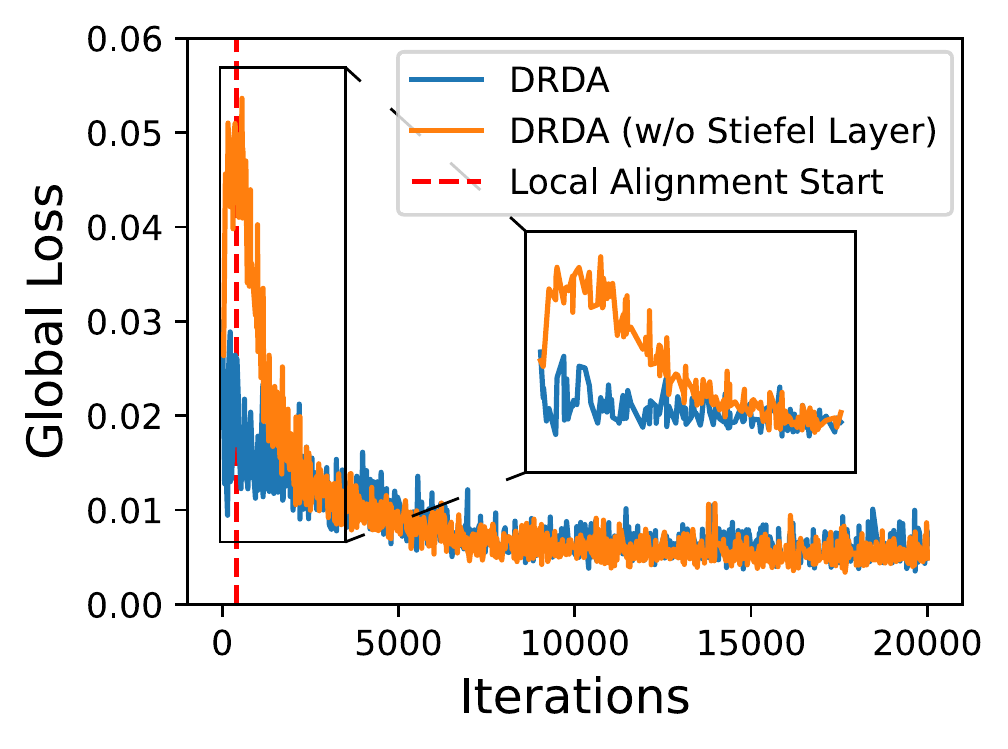}    \includegraphics[width=0.48\linewidth]{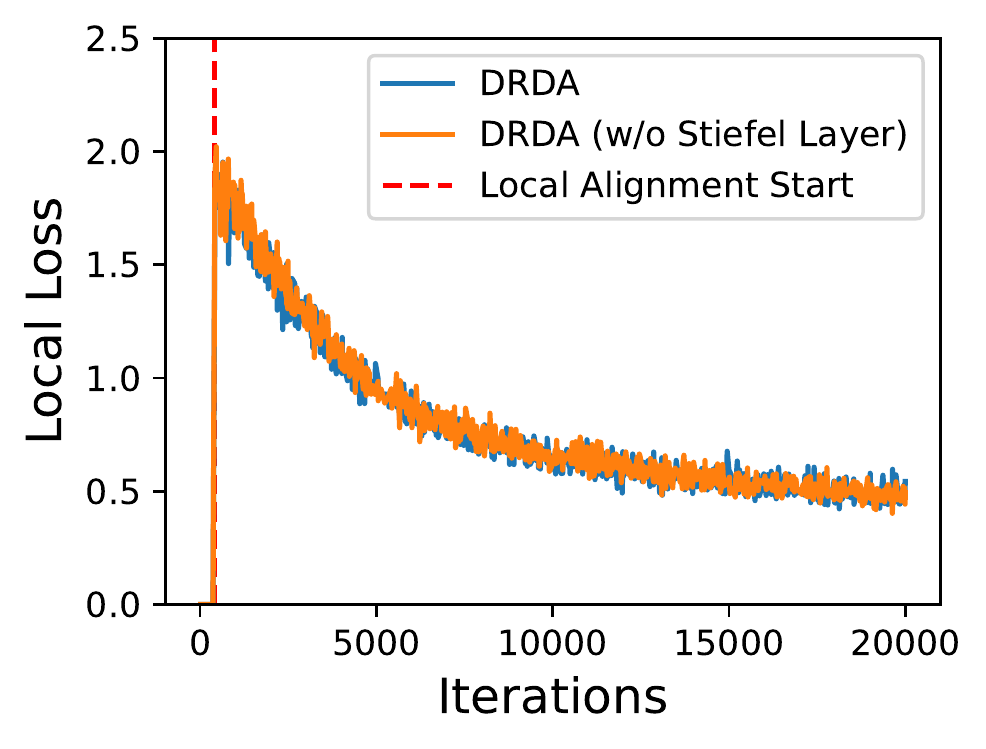}
    \caption{Global translation distance $\mathcal{L}_{\rm global}$ and local structure loss $\phi(\mathcal{V}^{s},\mathcal{V}^{t})$.}
    \label{fig:wo_stiefel_global_loss}
\end{figure}
\subsubsection{Low-dimensional radial structures}

To illustrate the convergence of the structure, we built a simplified LetNet similar to \cite{chenAdversarialLearnedLossDomain2020} while reducing the bottleneck dimension to 2 and training it with the task MNIST$\rightarrow$USPS.
We visualize the two dimensional features from different stages of the overall training procedure, with different colored points showing different categories.
To facilitate visual understanding, the source domain features and the target domain features were plotted side-by-side in separate plots, and instance points were uniformly sampled from entire datasets with stride 10.
In Fig.\ref{fig:structure_evolution}, we observe that the features coming from the source and target domains gradually evolve the radial-like structures.
It is worth noting that, at the beginning of the process, there is a significant structure discrepancy between the source and target.
Then, as the training progresses, the structures of two domains become more and more discriminative (i.e. the features in latent space present a more and more clear radial-like manner). Also, the decreasing of both source structure error $\phi(\mathcal{V}^{s},\mathcal{V}^{s}_{GT})$ and target structure error $\phi(\mathcal{V}^{t},\mathcal{V}^{t}_{GT})$ indicates these radial-like structures become more and more close to ground-truth ones.
As radial-like structures become more reliable and discriminative, domain alignment is expected to be more accurate as well.

\subsubsection{Global isometric effects}

To evaluate the proposed Stiefel layer, we implement a network without the Stiefel layer for comparison, denoting as DRDA (w/o Stiefel).
The detailed performance results presented in Table~\ref{tab:office-31},\ref{tab:office-home},\ref{tab:multi_domain},\ref{tab:domain_agnostic} indicate the importance of the Stiefel layer to the final accuracy.
Based on the results of the performance drop, we can confirm the hypothesis that features will be rotated globally and that the Stiefel layer can reduce the negative effect on domain alignment.
Furthermore, we tried to elaborate on the in-depth explanations in Fig.\ref{fig:wo_stiefel_global_loss}.
As we can see, when we use the Stiefel layer, we can reduce the structure loss to a small range quickly, and the maximum loss values are significantly less than those obtained without the Stiefel layer.
This is partly due to the fact that the global rotation difference between the source and target domains is particularly a misleading factor in the calculation of domain discrepancies in the early stages of domain alignment.
Furthermore, this misleading factor can have irreversible negative impacts on overall domain alignment. The final accuracy drops for the models without the Stiefel layer also confirmed these irreversible negative impacts.
The global rotational component can be easily extracted from the difference between two radial structures when there is a Stiefel layer, which would naturally mitigate such negative impacts in the early stages of domain alignment.
Additionally, the results of this study demonstrate the necessity of decoupling global and local transformations when performing alignment and the Stiefel layer is a suitable choice.

\subsubsection{Effects of structures alignment}

To better understand the effects of radial structure local alignment on domain alignment, we perform an ablation study that does not optimize for local structure loss at the beginning of the training period.
The detailed results are shown in Fig.\ref{fig:evolution_process}.
As we can see accuracy on target increasing during the early learning stage and decreases while training moves on.
As we can see obviously when ‘burn\_in:0.5’ the accuracy on target increases during the early learning stage and then gets stuck. Meanwhile, the structure loss, the Wasserstein distance from instances to local anchors, and the KL divergence between geometrical labels and classifier labels were increased.
It is clear from these simultaneous losses increasing that the structure of the target domain is crumbling.
This is because, with the progress of training, the network gradually learns the common semantic information at the beginning, and then begins to over-fit the data in the source domain.
Moreover, this over-fit phenomenon is accompanied by arbitrary distortions to discriminative structures. 
As shown in Fig.\ref{fig:evolution_process}, stretching the ‘burn\_in’ results in irreversible damage to the final accuracy. 
By comparing the influences of different ‘burn\_in’ on the rest component losses in Fig.\ref{fig:evolution_process}, we notice that once the structure alignment operations are restored, the corresponding losses drop rapidly.
Correspondingly, the test errors on the target domain also decreased rapidly after structure alignment was restored. 
The results show that our structure alignment can always reconstruct and align discriminative structures, which supports the validity of our model in the domain alignment.

\subsubsection{Angular term effects in intra-structure comparison}
When the angular losses are removed from the intra-structure comparison loss function, denoting as DRDA (w/o Angular), the performance returns to baseline, which indicates that discrepancy based on angular distance between discriminative vectors is very critical. The reason can be two folds. First, the angular loss is more consistent with the formulation of classification. Secondly, in high-dimensional space, the mass of the sphere is primarily concentrated on the shell and the distance between any two point pairs becomes even smaller. 
Therefore, angular loss confirms that modeling data distribution with radial-like structures that are well suited to angular comparison is an effective strategy.

\subsubsection{Effects of optimal transport distance minimization}
To verify the effectiveness of optimal transport distance minimization between instance and local anchors, we conduct the ablation studies by removing this minimization term in training, denoting as DRDA (w/o OT). The detailed results reported in Table~\ref{tab:office-31} and \ref{tab:office-home} illustrate the performance drop when instances are not restricted to being located nearby local anchors. We note that the performance degradation of DRDA (w/o OT) is much smaller than that of DRDA (w/o Angular), indicating that our proposed radial structure based alignment is efficient and robust in domain adaptation even if the structures are not forced to be compact.

\subsubsection{Effects of consensus regularization}
We evaluate the proposed classifier regularization terms by implementing the model without regularization loss $\mathcal{R}(\mathbf{P},\mathbf{Q})$ denoting as DRDA (w/o $\mathcal{R}_{\varphi}$). The results reported in Table~\ref{tab:office-31} and \ref{tab:office-home} indicate that such a regularization term enhances the performance of domain alignment across all subtasks. It is shown that the consensus regularization between geometrical assignments and classifier assignments can enhance the classification performance of the classifier by encouraging the classifier to admit geometrical assignments.

\subsubsection{Parameters sensitivity analysis}

In this section, we conduct sensitivity analysis on the hyper-parameters for our proposed method. The detailed results are shown in Fig.\ref{fig:param_sensitivity}. Parameters $\lambda_{T}$, $\lambda_{\phi}$, $\lambda_{ot}$ and $\lambda_{R}$ are mainly for scaling the loss value. 
From the observation, we choose $\lambda_{T}=150$ which controls the impacts of global translation between domains. The parameters $\lambda_{ot}$ balance the radial-like structure compactness and alignment effects, and we find when $\lambda_{ot}$ is around 0.0001-0.005 the model performance reaches the peaks. The parameters $\lambda_{R}$ regularize the agreement between classifier and geometric assignments, when $\lambda_{R}=20.0$ the model performance reaches the peaks. As for structure alignment, $\lambda_{\phi}=3$ provides the best performance.
It is also obvious that the performance of the system is quite stable across a wide range of transfer losses when they are ranged in respective orders of  magnitude.
\begin{figure}[ht]
    \centering
    \includegraphics[width=\linewidth]{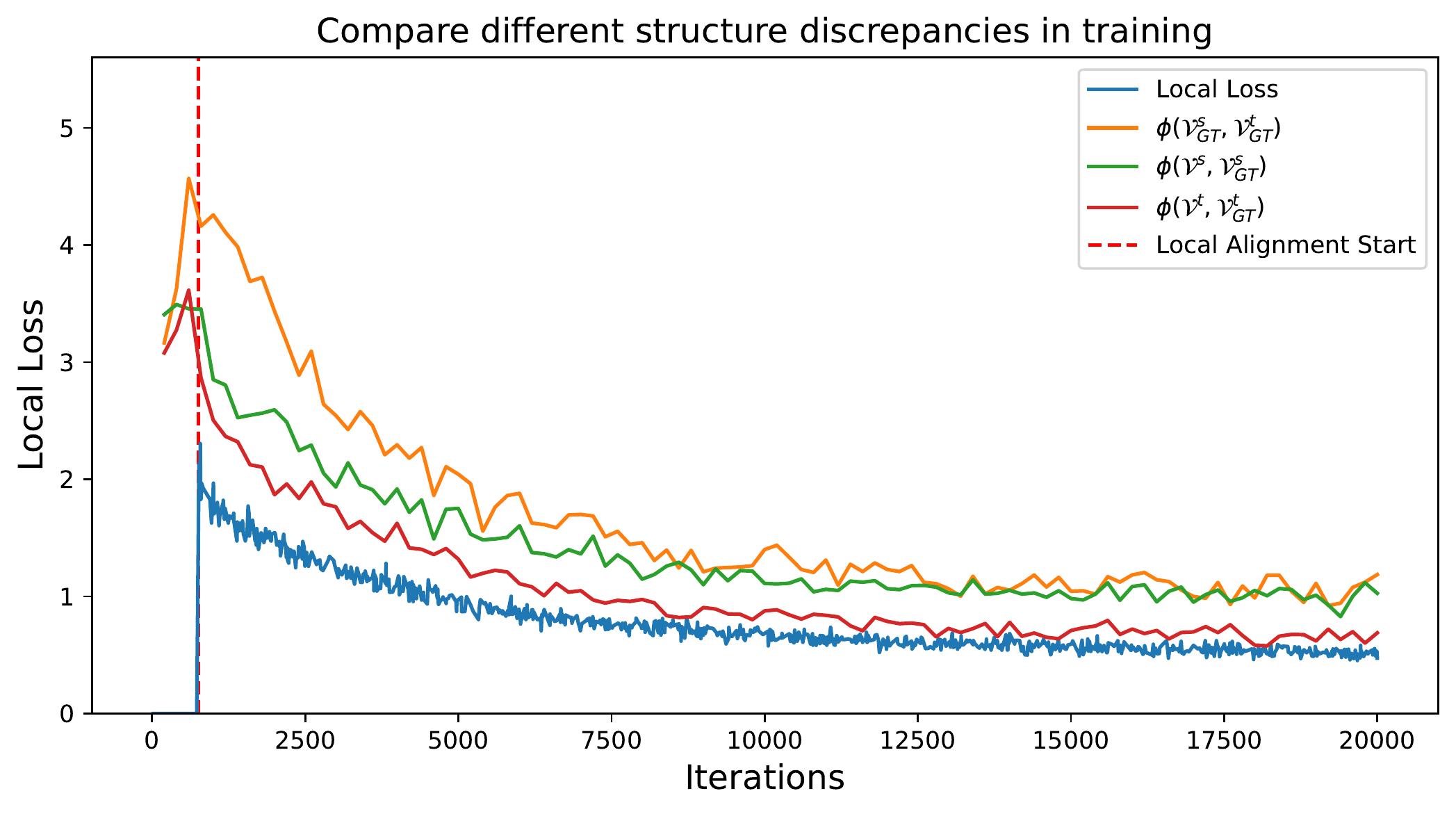}
    \caption{Radial structures alignment convergence visualization with an increasing number of iterations, where $\phi(\mathcal{V}^{s}$,$\mathcal{V}^{s}_{GT})$,$\phi(\mathcal{V}^{t},\mathcal{V}^{t}_{GT}),\phi(\mathcal{V}^{s}_{GT},\mathcal{V}^{t}_{GT})$ are computed every validation, $\mathcal{V}^{s}_{GT}$ indicates the structure calculated with ground truth labels (resp $\mathcal{V}^{t}_{GT}$).}
    \label{fig:structure_diff}
\end{figure}
\subsubsection{The evolution of structures discrepancies}
We also demonstrate the differences of the radial structure between domains with increasing iteration numbers.
As illustrated in Fig.\ref{fig:structure_diff}, after adding the structure alignment, the discrepancies $\phi(\mathcal{V}^{s}_{GT},\mathcal{V}^{t}_{GT})$ of structures derived from ground truth labels appear to consistently decrease, as well as local differences between structures being minimized. It is evident from the results that our method is able to consistently produce positive alignment with the increasing number of training iterations.

\section{Conclusion}
This paper presents a new structure-preserved domain adaptation method, which has two key features: a new discriminative radial structure and a new alignment strategy based on radial structure. 
The discriminative radial structure preserves both representative and discriminative information in feature distribution.
The decoupled global alignment and fine-grained morphological alignment reduce the common domain shifts and conditional domain shifts.
Experimental results on several benchmark datasets showed that 
i) our method consistently outperforms state-of-the-art methods on four types of unsupervised domain adaptation tasks, and 
ii) our method leads to more superiority when the task is more challenging.



\section*{Acknowledgment}
This work is supported by the National Key R\&D Program of China(2020YFB1313501), Zhejiang Provincial Natural Science Foundation (LR19F020005) , National Natural Science Foundation of China (61972347, T2293723) and the Fundamental Research Funds for the Central Universities (No. 226-2022-00051).

\ifCLASSOPTIONcaptionsoff
  \newpage
\fi



\bibliographystyle{IEEEtran}
\bibliography{references}

%

\begin{IEEEbiography}[{\includegraphics[width=1in,height=1.25in,clip,keepaspectratio]{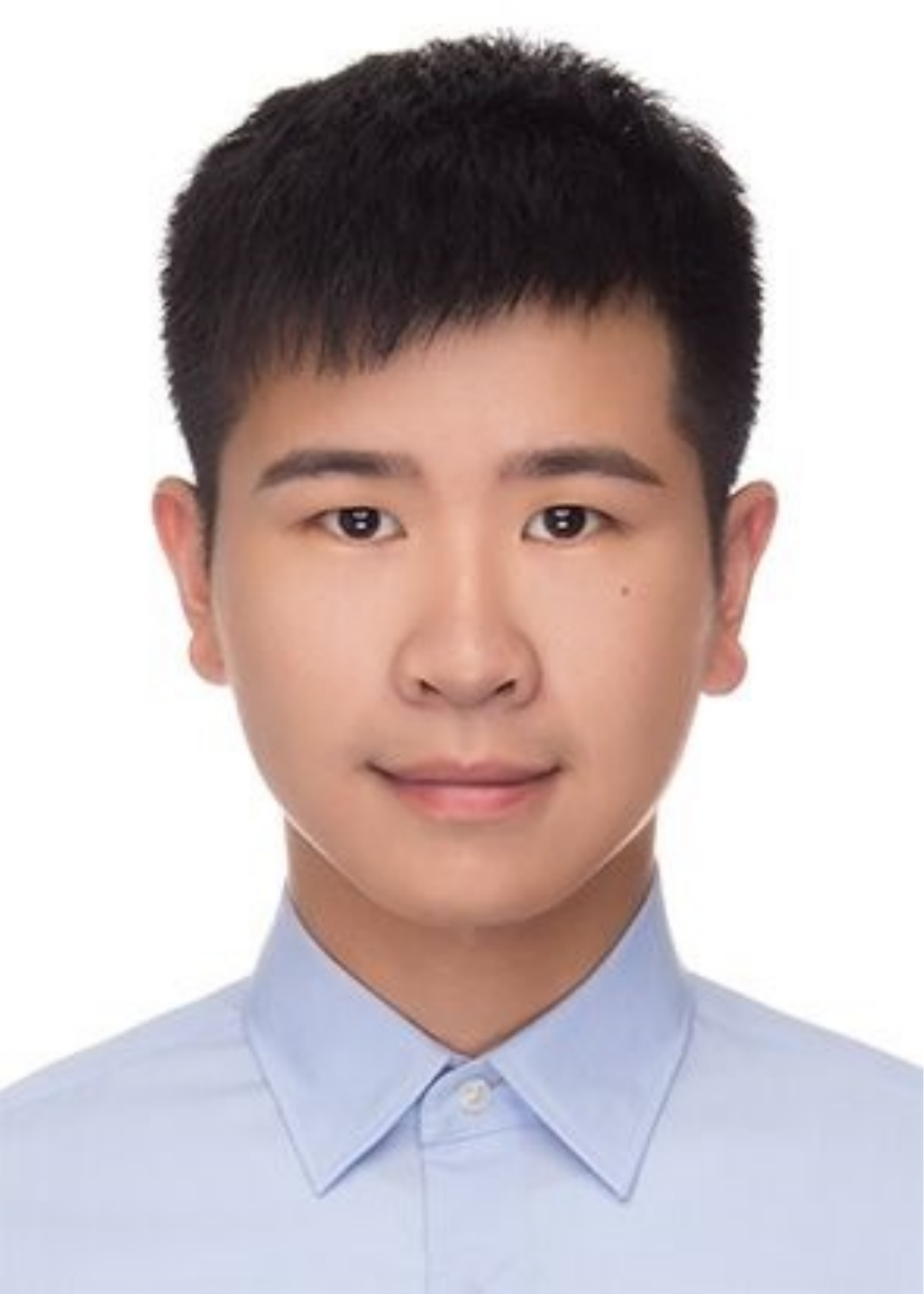}}]{Zenan Huang} received B.E. degree in the Computer Science from Zhejiang University of Technology, in 2018. He is currently pursuing the Ph.D degree in the College of Computer Science and Technology, Zhejiang University. His research interests include computer vision, causalility, and machine learning.
\end{IEEEbiography}

\begin{IEEEbiography}[{\includegraphics[width=1in,height=1.25in,clip,keepaspectratio]{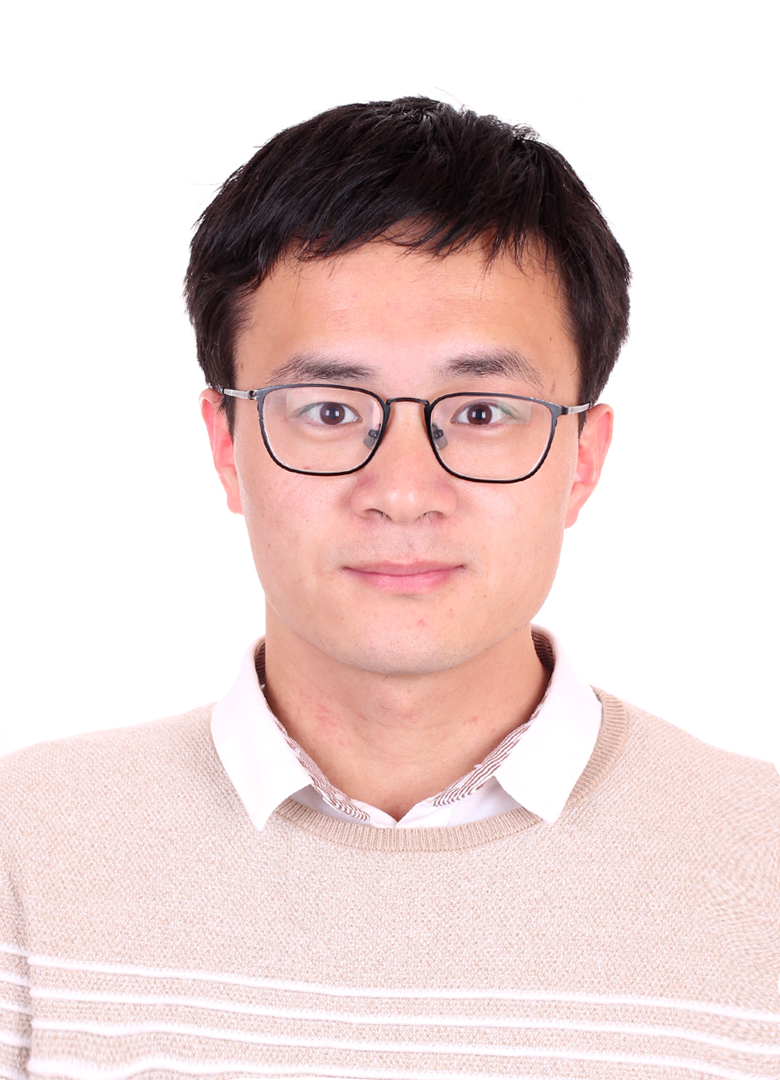}}]{Jun Wen}
  received the Ph.D. degree in computer science from Zhejiang University, Hangzhou, China, in 2020. He is currently a Postdoctoral Research Fellow at the Harvard Medical School. His research interests include transfer learning and biomedical informatics.
\end{IEEEbiography}


\begin{IEEEbiography}[{\includegraphics[width=1in,height=1.25in,clip,keepaspectratio]{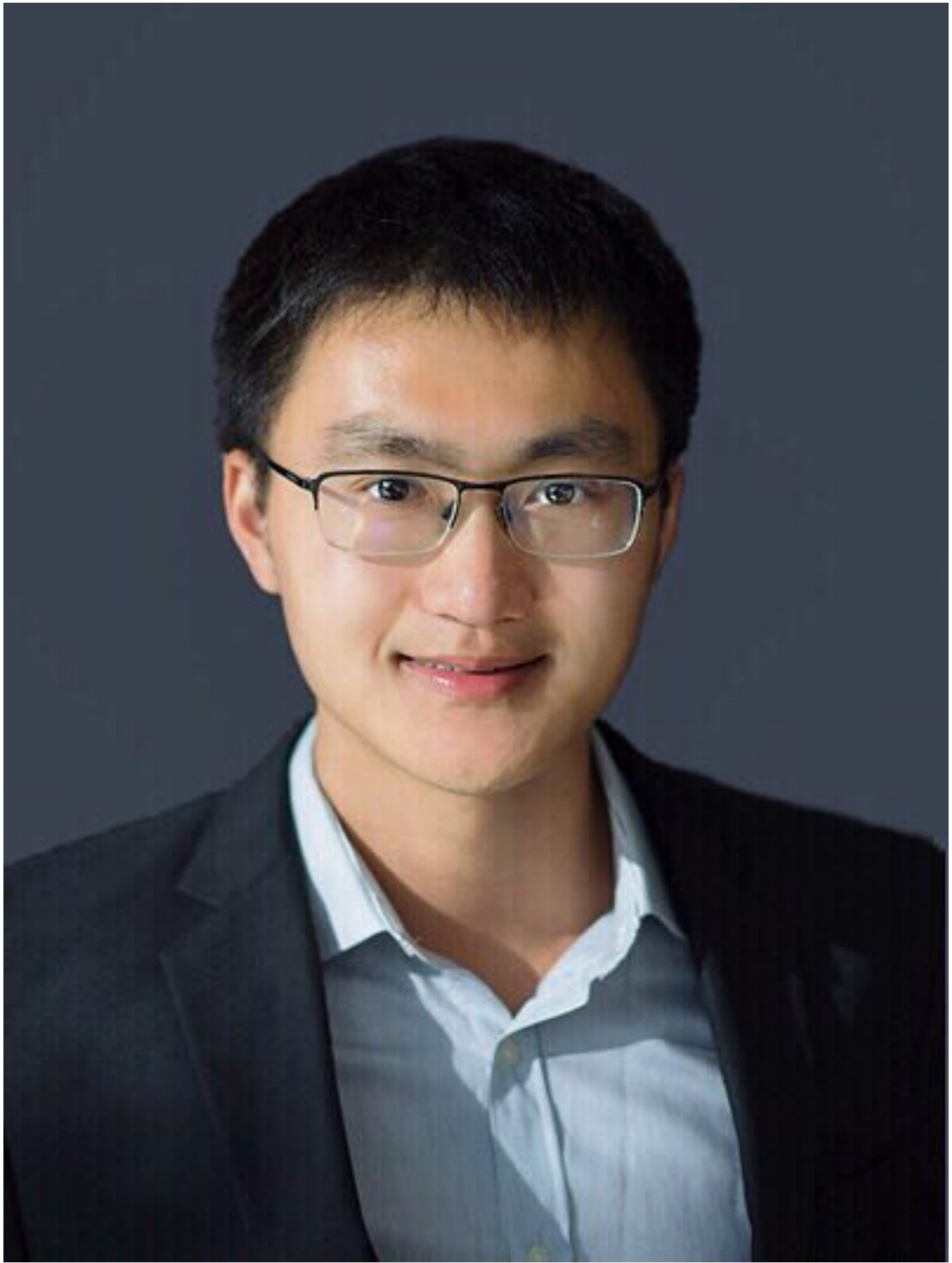}}]{Siheng Chen}  is a tenure-track associate professor of Shanghai Jiao Tong University and Co-PI at Shanghai AI Laboratory. Dr. Chen received his doctorate from Carnegie Mellon University. Dr. Chen's work on sampling theory of graph data received the 2018 IEEE Signal Processing Society Young Author Best Paper Award. His co-authored paper on structural health monitoring received ASME SHM/NDE 2020 Best Journal Paper Runner-Up Award and another paper on 3D point cloud processing received the Best Student Paper Award at 2018 IEEE Global Conference on Signal and Information Processing. Dr. Chen contributed to the project of scene-aware interaction, winning MERL President's Award. His research interests include collective intelligence, autonomous driving and graph neural networks.
\end{IEEEbiography}

\begin{IEEEbiography}[{\includegraphics[width=1in,height=1.25in,clip,keepaspectratio]{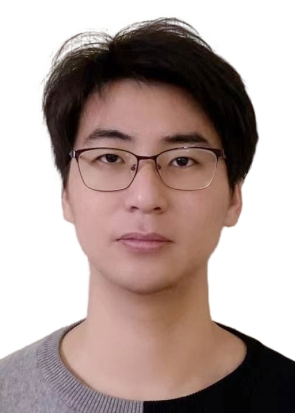}}]{Linchao Zhu} (Member, IEEE) received the B.E. degree from Zhejiang University, China, in 2015, and the Ph.D. degree in computer science from the University of Technology Sydney, Australia, in 2019. He is a Research Professor with the College of Computer Science and Technology, Zhejiang University, China. His research interests are video analysis and understanding.
\end{IEEEbiography}

\begin{IEEEbiography}[{\includegraphics[width=1in,height=1.25in,clip,keepaspectratio]{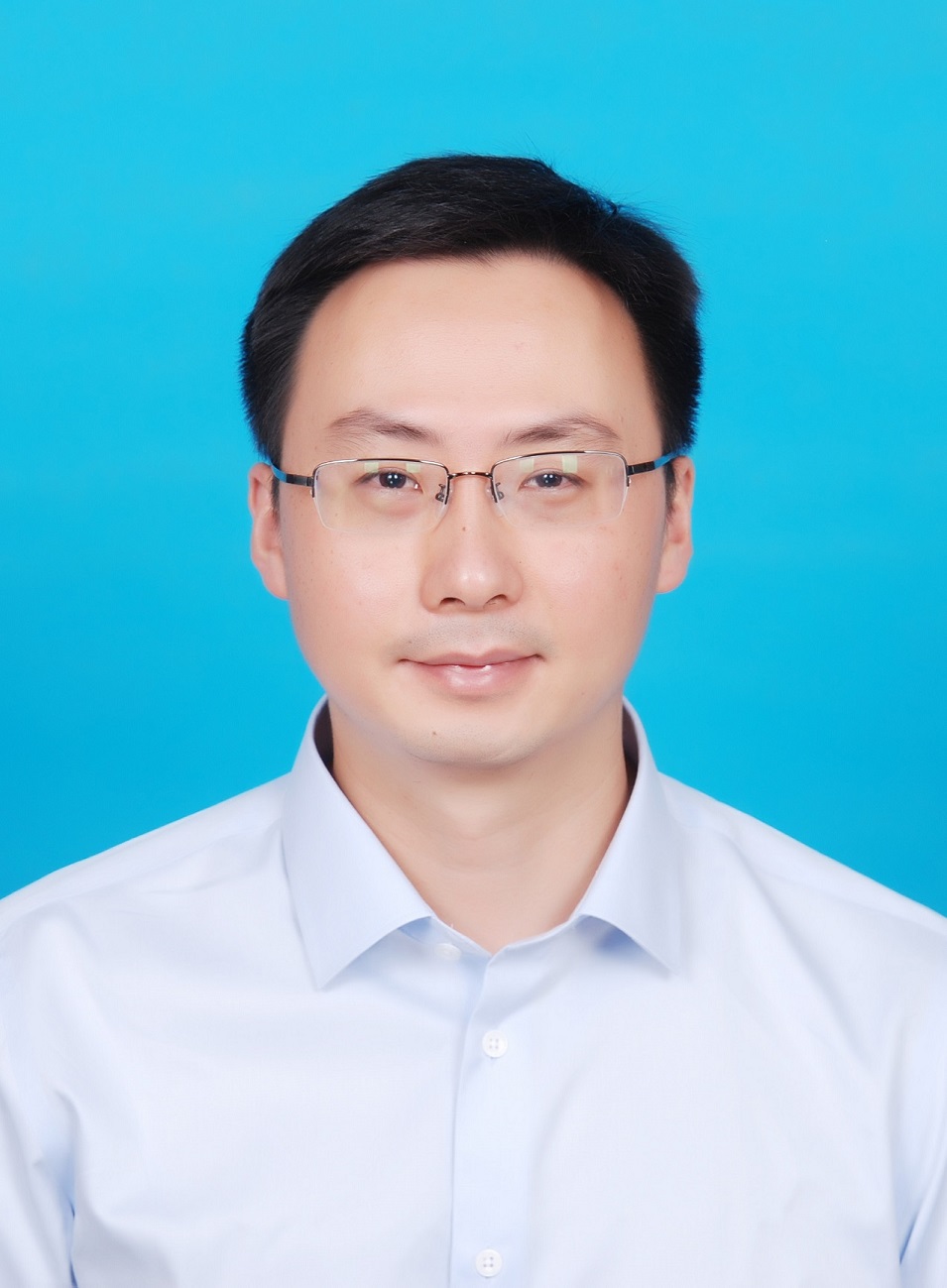}}]{Nenggan Zheng} received the bachelor's and Ph.D. degrees from Zhejiang University, Hangzhou, China, in 2002 and 2009, respectively. He is currently a Full Professor in computer science with the Academy for Advanced Studies, Zhejiang University. His research interests include artificial intelligence, brain–computer interface, and embedded systems.
\end{IEEEbiography}


\vfill


\end{document}